\documentclass[11pt,letterpaper]{article}
\usepackage[utf8]{inputenc}
\usepackage[T1]{fontenc}
\usepackage[english]{babel}
\usepackage[margin=0.8in]{geometry}
\usepackage{newtxtext,newtxmath}
\usepackage{microtype}
\usepackage{amsmath}
\usepackage{graphicx,xcolor}
\usepackage{booktabs,multirow,makecell,threeparttable,colortbl}
\usepackage{pifont,inconsolata}
\usepackage{authblk}
\usepackage{caption}
\usepackage{titlesec}
\usepackage[numbers,sort&compress]{natbib}
\usepackage{xurl}
\usepackage[hidelinks]{hyperref}
\usepackage{bookmark}
\titleformat{\section}{\large\bfseries}{\thesection}{0.6em}{}
\titleformat{\subsection}{\normalsize\bfseries}{\thesubsection}{0.6em}{}
\titlespacing*{\section}{0pt}{2.5ex plus 1ex minus .2ex}{1ex}
\titlespacing*{\subsection}{0pt}{2ex plus .5ex minus .2ex}{0.5ex}
\newcommand{\methodname}{SyRe}
\newcommand{\datasetname}{SyReData}

\hypersetup{pdftitle={Synergistic Vision-Language Reinforcement Enables Scalable On-Demand Analysis across Diverse Clinical Tasks},pdfauthor={Haonan Wang and collaborators},pdfsubject={SyRe: text-prompted medical image segmentation}}
\title{\LARGE\bfseries Synergistic Vision-Language Reinforcement Enables Scalable On-Demand Analysis across Diverse Clinical Tasks}
\author[1,$\dag$]{Haonan Wang}
\author[2,$\dag$]{Jiaji Mao}
\author[1]{Lehan Wang}
\author[1]{Qixiang Zhang}
\author[1]{Marawan Elbatel}
\author[1]{Yi Qin}
\author[2]{Huijun Hu}
\author[2]{Baoxun Li}
\author[2]{Wenhui Deng}
\author[2]{Weifeng Qin}
\author[1]{Hongrui Li}
\author[1]{Jialin Liang}
\author[2,*]{Jun Shen}
\author[1,3,*]{Xiaomeng Li}
\affil[1]{Department of Electronic and Computer Engineering, The Hong Kong University of Science and Technology, Hong Kong SAR}
\affil[2]{Department of Radiology, Guangdong Provincial Key Laboratory of Malignant Tumor Epigenetics and Gene Regulation, Sun Yat-Sen Memorial Hospital, Sun Yat-Sen University, Guangzhou 510120, China}
\affil[3]{Department of Computer Science and Engineering, The Hong Kong University of Science and Technology, Hong Kong SAR}

\affil[*]{Corresponding authors: Xiaomeng Li (eexmli@ust.hk); Jun Shen (shenjun@mail.sysu.edu.cn)}
\affil[$\dag$]{These authors contributed equally to this work}
\date{}
\begin{document}
\maketitle
\begin{abstract}
Accurate delineation of tumors and surrounding organs-at-risk in medical images is fundamental to modern oncology, guiding radiotherapy, surgery and treatment response assessment. Yet this process remains time-consuming and expertise-intensive, and current artificial intelligence (AI) systems are not a complete solution: they typically require manual spatial prompts or costly task-specific retraining, and often rely on generic class labels that provide weak semantic grounding for heterogeneous disease targets. As a result, a single on-demand tool that can support segmentation-dependent clinical analyses through simple text instructions has remained elusive.
Here we present SyRe, a promptable segmentation foundation model based on \textbf{Sy}nergistic vision-language \textbf{Re}inforcement. Rather than using language as a shallow prompt, SyRe strengthens bidirectional interaction between visual and linguistic representations to improve semantically grounded spatial understanding. To enable this at scale, we introduce the Color Region Description (CRD) strategy and construct \datasetname{}, a large-scale language-driven segmentation dataset containing 20 million image-mask-description triplets across 9 modalities and 229 segmentation tasks, and train SyRe under diversified prompt forms rather than a single fixed query per image.
SyRe achieves accurate text-prompted segmentation across diverse clinical scenarios, with especially strong gains on disease-related targets. Together, these design choices enable open-ended prompting, invalid-prompt rejection and flexible switching between single-target and multi-target analysis. Across 28 unseen external datasets, including 17 cancer types, 3 non-cancer diseases and multinational in-house cohorts, SyRe generalizes robustly under real-world distribution shift. Beyond spatial overlap, SyRe-generated masks preserve clinically relevant quantitative information in pathology and, across five retrospective tumor imaging cohorts spanning CT and MRI, yield radiomics features that stratify survival and improve prognostic modeling over clinical baselines. Finally, clinician-in-the-loop refinement supports efficient case-level correction when greater precision is required. Together, these results establish SyRe as a generalizable foundation for scalable quantitative oncology analysis and clinician-guided segmentation refinement.
\end{abstract}

\section{Introduction}

Reliable interpretation of medical images is central to modern healthcare, often determining whether patients receive timely diagnosis, appropriate treatment, and improved survival~\cite{beauchamp2023integrative}. 
Among medical imaging tasks, precise segmentation of organs, tumors and pathological lesions is especially important because it directly supports surgical planning, radiotherapy, longitudinal disease monitoring and quantitative assessment of treatment response~\cite{UNet_2015,oktay2018attentionunet,isensee2021nnunet,cao2022swin,SEG_wang2022uctransnet,SemiSEG_AND,SemiSEG_DHC,natseg_chen2023scs,natseg_cutler2022omnipose,natseg_pang2024cellotype,NMI_peiris2023uncertainty,NMI_sekh2021physics}. 
However, accurate image interpretation remains highly specialized. Different modalities, including computed tomography (CT), magnetic resonance imaging (MRI), X-ray, pathology, ultrasound, fundus and endoscopy, require domain-specific expertise, and such expertise is often scarce in under-resourced regions and high-burden clinical settings.

This annotation bottleneck is especially limiting for retrospective clinical research. 
Large oncology cohorts often contain hundreds or thousands of imaging studies linked to survival, recurrence, or treatment-response outcomes, but extracting quantitative tumor morphology from such cohorts requires expert delineation of every lesion. 
Although clinicians can perform this task for selected cases, doing so across an entire retrospective cohort can require hundreds of hours of specialist time, making many potentially valuable analyses impractical in routine clinical departments. 
A clinically useful on-demand image analysis system should therefore not only segment individual cases accurately, but also enable scalable extraction of quantitative imaging biomarkers from large cohorts without exhaustive manual delineation or task-specific model development.

Recent advances in multimodal foundation models have raised the prospect of a universal medical AI assistant that could support clinicians across institutions and tasks~\cite{huang2023visual,natmed_VLM_lu2024visual,natmedVLM_christensen2024vision,nat_fu2025foundation,nat_hollmann2025accurate,nat_xiang2025vision,natFM_zhao2024foundation,natbmeFM_bluethgen2024vision,natbmeFM_sun2024foundation,natmedFM_zhang2024generalist,cui2024scgpt,ma2024multimodality,ma2023towards,NMI_pai2024foundation,NMI_perez2025exploring}. 
Yet current systems remain limited when fine-grained spatial understanding is required. 
Many vision-language medical assistants can answer questions or generate reports, but they cannot reliably delineate the exact boundaries of clinically meaningful structures. 
Conversely, strong segmentation models often require manual point or box prompts from expert users~\cite{ma2024medsam}, or task-specific retraining and engineering effort for each downstream application~\cite{isensee2021nnunet}. 
These requirements are particularly problematic for large retrospective cohorts, where manually specifying spatial prompts or developing task-specific models for every tumor type and modality can offset the intended efficiency gains. 
As a result, existing approaches fall short of a clinically practical paradigm in which a single model can respond to diverse text instructions and return spatially precise outputs on demand. This limitation is particularly consequential for retrospective oncology research, where large imaging cohorts may contain survival or treatment-response outcomes, but quantitative tumor analysis remains constrained by the need to manually delineate every lesion.

A key reason for this gap is weak semantic grounding. 
In current language-driven segmentation systems, text is often reduced to a generic class label, such as ``liver tumor'' or ``lung lesion''. 
However, disease-related targets are intrinsically heterogeneous: lesions within the same category can vary substantially in shape, size, texture and location across patients and modalities. 
Such generic labels therefore provide weak supervision for aligning language with visual regions, particularly for clinically important abnormalities. 
We hypothesized that overcoming this bottleneck requires richer region-level semantic supervision, so that image, mask and language representations are aligned not only at the category level, but also at the level of morphology and spatial context.
A second, less examined bottleneck concerns how language is used during training.
Existing language-driven segmentation models are typically trained with a single fixed query form, in which one named target is requested per image. Such training provides no incentive to distinguish which structures a given prompt does and does not refer to, and consequently offers little support for the open-ended, multi-target or ill-posed queries that arise in realistic clinical use.

Here we present \methodname{}, a unified framework for promptable medical image analysis based on \textit{synergistic vision-language reinforcement}. 
Rather than treating language as a shallow prompt attached to a segmentation model, \methodname{} explicitly reinforces bidirectional interaction between visual and linguistic representations: visual cues improve language understanding of the target, while textual semantics guide spatial localization and delineation. 
To enable this at scale, we introduce the \textit{Color Region Description} (CRD) strategy, which converts segmentation masks into color-coded regions and uses an off-the-shelf vision-language model to generate descriptions of their shape and relative position. 
To address the second bottleneck, we further introduce \textit{prompt-diversified training} (PDT). During training, PDT interleaves three formats — \textit{seg-all with CRD}, \textit{seg-all} without CRD, and \textit{seg-subset} — so that the model learns prompt-conditioned rather than image-conditioned behavior. A single-target request is the special case of \textit{seg-subset} in which the sampled subset contains one target, rather than an independent training branch. At test time, the model is queried with a separate protocol that supports single-target, multi-target and open-ended \textit{seg-all} requests.
Using this strategy, we construct \datasetname{}, a large-scale language-driven segmentation dataset containing 20 million image-mask-description triplets across 9 modalities and 229 segmentation tasks, substantially expanding the scale and diversity of semantically enriched supervision for medical segmentation.

This combination of richer supervision and unified multimodal modeling enables a different operating paradigm from prior work. Instead of requiring task-specific retraining or expert-crafted spatial prompts, \methodname{} supports accurate text-prompted segmentation across diverse clinical scenarios and generalizes robustly across unseen centers, populations, and tumor types. More importantly, across five retrospective tumor imaging cohorts spanning CT and MR, radiomics features extracted from \methodname{}-generated masks stratify survival and improve prognostic modeling over clinical baselines. \methodname{} further supports clinician-guided refinement when individual cases require greater segmentation precision. Together, these capabilities position \methodname{} as a foundation for scalable quantitative oncology analysis and interactive segmentation support rather than autonomous diagnosis.

\section{Results}

\begin{figure}[p]
\centering
\includegraphics[width=\linewidth]{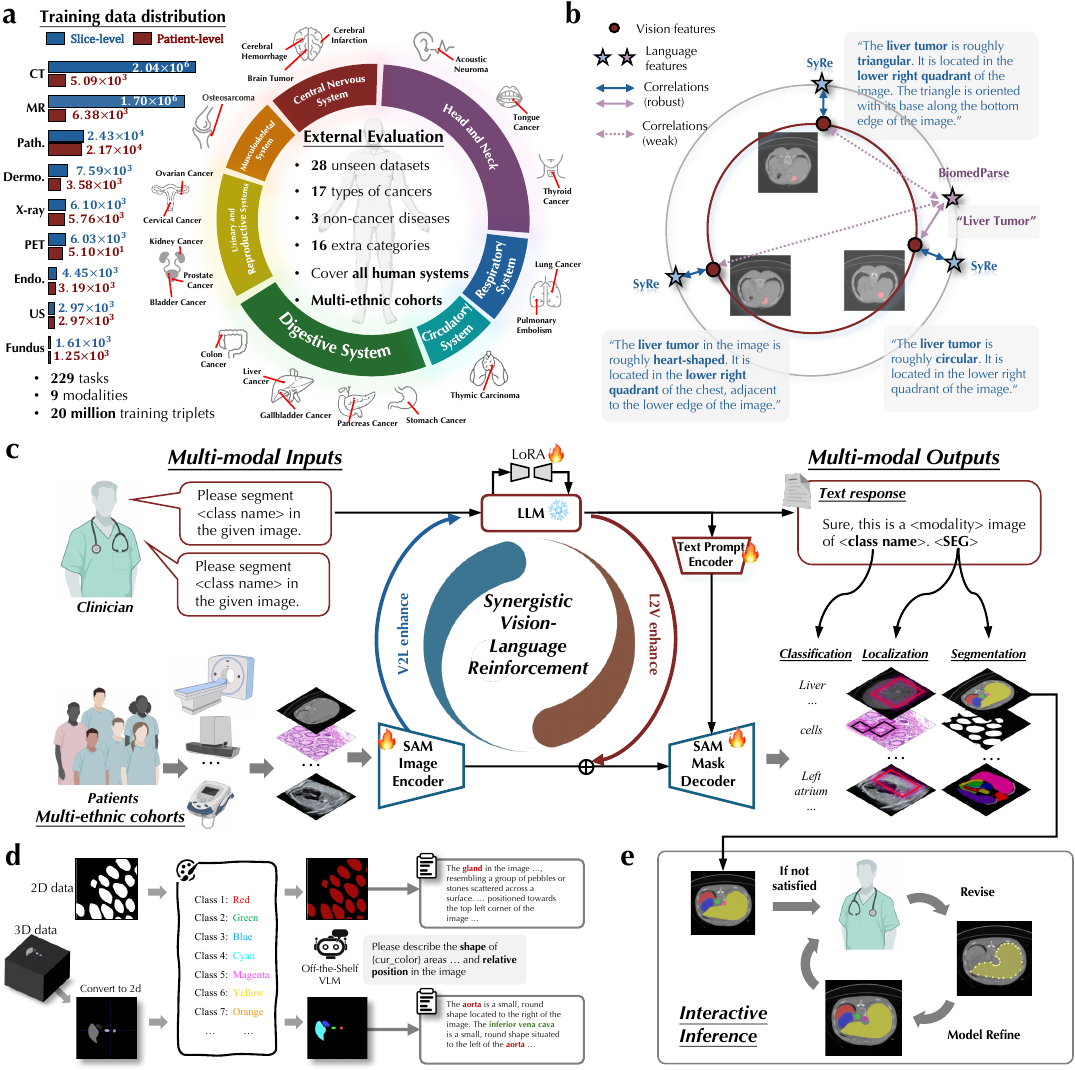}
\caption{
\textbf{Overview of \methodname{} and \datasetname{}.}
\textbf{a,} Bar plots show the slice-level and patient-level data distributions across modalities. The compact axis labels denote Path. (pathology), Dermo. (dermoscopy), Endo. (endoscopy), and US (ultrasound). The external evaluation set comprises 28 unseen datasets, spanning 17 cancer types, 3 non-cancer diseases (cerebral haemorrhage, cerebral infarction and pulmonary embolism) and 16 additional non-malignant, anatomical or procedural categories\ across major human systems.
\textbf{b,} Motivation for semantic region description. A generic class label such as ``liver tumor'' can correspond to lesions with diverse shapes and spatial locations, providing weak semantic grounding for image-text alignment.
\textbf{c,} Flowchart of \methodname{}. The framework takes multimodal medical images and text prompts as input, and produces text responses, classification, localization and segmentation outputs.
\textbf{d,} Pipeline of the Color Region Description (CRD) strategy. Segmentation masks from 2D images and 3D volumes are converted into colored region maps, which are then described by InternVL-1.5 to produce region-level semantic supervision.
\textbf{e,} \methodname{} supports interactive refinement during inference.
}
\label{fig:main_1}
\end{figure}

\begin{figure}[p]
\centering
\includegraphics[width=0.88\linewidth]{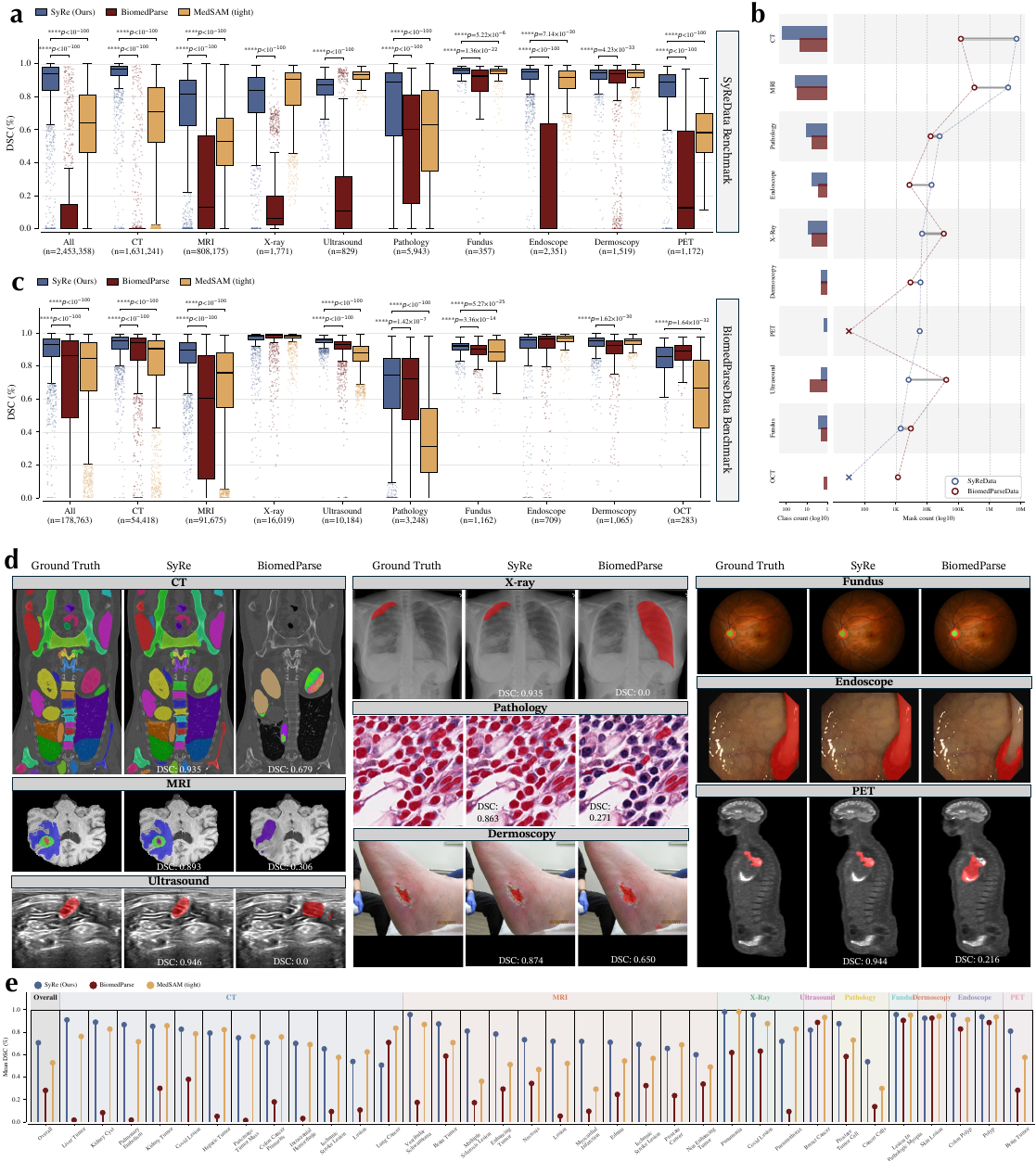}
\caption{
\textbf{\methodname{} shifts heterogeneous disease-target segmentation into the high-accuracy regime across modalities.}
\textbf{a,} Comparison on SyReData Bench, shown overall and across modalities. 
Both \methodname{} and BiomedParse use text prompts, and \methodname{} achieves higher segmentation accuracy without requiring manually specified spatial prompts. MedSAM (tight), by contrast, relies on manually specified disease-region prompts. P values indicate the significance of the difference between \methodname{} and the best competing method (two-sided paired t-test).
\textbf{b,} Comparison of benchmark complexity between SyReData Bench and BiomedParseData Bench across ten modalities. SyReData Bench contains more object categories and more masks per category for most modalities, indicating greater semantic diversity and higher task difficulty.
\textbf{c,} Comparison on BiomedParseData Bench. \methodname{} achieves the best overall performance and maintains strong results across modalities, showing that its advantage generalizes beyond the more challenging SyReData Bench.
\textbf{d,} Representative qualitative comparisons on internal datasets. \methodname{} produces masks that more closely match the ground truth than BiomedParse.
\textbf{e,} Category-wise performance on disease classes across ten modalities. \methodname{} outperforms competing methods on most categories, with especially clear gains for heterogeneous lesion and tumor classes.
}
\label{fig:main_2}
\end{figure}

\subsection{Morphology- and spatial-context-grounded supervision for text-prompted medical segmentation}

We first established a large-scale heterogeneous training and evaluation framework for language-driven medical image segmentation. 
\datasetname{} comprises 20 million image-mask-description triplets spanning 9 imaging modalities and 229 segmentation tasks, covering a broad range of anatomies and disease categories. Here, a “segmentation task” is one task-level benchmark unit defined by a target concept together with its imaging/data context; it is distinct from the number of canonicalized unique class labels or the number of rows used in a multi-benchmark coverage table.
To assess real-world generalizability, we further constructed a broad external benchmark including 28 unseen datasets, 17 cancer types, 3 non-cancer diseases (cerebral haemorrhage, cerebral infarction and pulmonary embolism) and 16 additional non-malignant, anatomical or procedural categories, \ together with multinational in-house validation data (Fig.~\ref{fig:main_1}a).

A central motivation for our framework is that generic class labels often provide weak semantic grounding for medical image understanding. This limitation is especially severe for disease-related targets, whose morphology and spatial location can vary substantially even within the same category. For example, a label such as ``liver tumor'' may correspond to lesions with markedly different shapes, sizes and positions, making it difficult to establish robust image-text correspondence from category names alone (Fig.~\ref{fig:main_1}b).

Based on this motivation, we developed \methodname{}, a unified multimodal framework for promptable medical image analysis. 
\methodname{} takes medical images and text prompts as input, and produces text responses, classification, localization and segmentation outputs within a single framework. 
The model introduces \textit{Synergistic Vision-Language Reinforcement}, a supervised vision-language mechanism in which visual and linguistic representations mutually enhance one another rather than using reinforcement learning, thereby improving semantic alignment for downstream medical image analysis (Fig.~\ref{fig:main_1}c).
To ensure that this alignment is robust to the way clinicians actually query the model, we further adopt \textit{prompt-diversified training} (PDT). During training, PDT interleaves \textit{seg-all with CRD}, \textit{seg-all} without CRD and \textit{seg-subset} formats rather than a single fixed target query. A single-target request is the $k=1$ case of \textit{seg-subset}, not a separate sampling branch. Testing uses a separate prompt protocol: the released model supports single-target, multi-target and open-ended \textit{seg-all} requests.

To provide semantically enriched supervision at scale, we further propose the \textit{Color Region Description} (CRD) strategy. 
CRD converts segmentation masks into color-coded regions and queries an off-the-shelf vision-language model to generate descriptions of the shape and relative position of each region (Fig.~\ref{fig:main_1}d and Extended Data Fig.~\ref{fig:extended_1}). 
Compared with generic class labels, these generated descriptions provide substantially richer semantic anchors for image-mask-text alignment, particularly for heterogeneous disease targets, and enable the construction of \datasetname{} as a large-scale language-driven segmentation dataset.

During training, CRD-generated descriptions are used to enrich semantic supervision and strengthen region-language alignment, whereas during standard inference, \methodname{} operates with text prompts alone and does not require access to segmentation masks or description guidance. 
The framework is also designed to support iterative clinician-in-the-loop refinement, allowing the model output to be revised when the initial prediction is not satisfactory and thereby providing a unified interface for on-demand medical image analysis in realistic clinical workflows (Fig.~\ref{fig:main_1}e).

\subsection{\methodname{} shifts heterogeneous disease-target segmentation into the high-accuracy regime across modalities}

We next evaluated whether \methodname{} can accurately segment clinically meaningful targets across diverse medical imaging scenarios using natural-language prompts alone. 
On the held-out \textit{SyReData} benchmark, which contains 2,453,358 samples spanning CT, MRI, X-ray, ultrasound, pathology, fundus, endoscopy, dermoscopy and PET, \methodname{} consistently achieved the best overall performance across modalities and substantially outperformed BiomedParse, while also surpassing or matching MedSAM supplied with tight box prompts in most settings (Fig.~\ref{fig:main_2}a). 
The class-wise results (Extended Data Fig.~\ref{fig:extended_3}) show that \methodname{} achieves DSC $> 0.8$ in 192 of the 229 evaluated class entries (83.8\%), compared with 58 class entries (25.3\%) for MedSAM and only 9 class entries (3.9\%) for BiomedParse. Under the stricter criterion of DSC $> 0.9$, \methodname{} still leads by a large margin, reaching 108 class entries (47.2\%), versus 22 (9.6\%) for MedSAM and 2 (0.9\%) for BiomedParse. 
These full-benchmark counts combine semantic vocabulary coverage with segmentation accuracy. To separate these effects, we re-analysed the class-level results by BiomedParse vocabulary coverage (Extended Data Fig.~\ref{fig:extended_8}). Among the 229 evaluated entries, 50 were shared with the BiomedParse inventory and 179 were unsupported. On the shared subset, mean DSC was 0.878 for \methodname{}, 0.384 for BiomedParse and 0.739 for MedSAM (tight), with DSC $>0.8$ in 42, 6 and 22 entries, respectively. On unsupported entries, BiomedParse cannot provide a like-for-like target-specific prediction; the corresponding full-benchmark values therefore quantify practical coverage as well as segmentation performance. Within this qualification, the results indicate that the advantage of \methodname{} is not limited to improving average performance on a small subset of favorable tasks, but extends across a broad set of evaluated anatomies, lesions and disease-related targets.
The performance gains were particularly evident in modalities where clinically relevant targets are highly heterogeneous, such as CT, MRI, pathology and PET, indicating that \methodname{} is able to robustly align text prompts with complex visual patterns rather than relying on narrow prompt-specific cues.

To place these results in context, we compared the complexity of the \textit{SyReData} and \textit{BiomedParseData} benchmarks (Fig.~\ref{fig:main_2}b and Extended Data Fig.~\ref{fig:extended_2} for class-wise comparison). 
SyReData contains substantially larger class and mask counts across major modalities, especially CT, MRI and pathology, making it a more challenging and clinically representative benchmark than BiomedParseData. 
Despite this increased difficulty, \methodname{} maintained strong performance across the full benchmark, suggesting that the gains do not arise from easier task composition but instead reflect improved robustness under large-scale heterogeneous clinical settings.

We further evaluated \methodname{} on the held-out \textit{BiomedParseData} benchmark, which contains 178,763 samples across CT, MRI, X-ray, ultrasound, pathology, fundus, endoscopy, dermoscopy and OCT (Fig.~\ref{fig:main_2}c). 
\methodname{} remained highly competitive on this independent benchmark and again consistently exceeded BiomedParse across modalities, while performing comparably to or better than MedSAM in most cases. 
The performance margins were smaller in dermoscopy and fundus, where the publicly available datasets are relatively limited and the visual patterns are more homogeneous, leaving less room for separation among already strong promptable models. 
These results indicate that the advantages of \methodname{} are not restricted to our own benchmark, but transfer to an existing large-scale medical segmentation benchmark with a different task composition and data distribution.

Qualitative comparisons on representative internal cases further highlight these advantages (Fig.~\ref{fig:main_2}d). 
Across diverse examples including pathological nuclei, brain lesions, spinal structures, abdominal organs and pelvic anatomy, \methodname{} produced masks that more closely matched the ground truth, whereas BiomedParse frequently missed target regions, confused adjacent structures or failed to recover complete object extent. 
These examples illustrate that the quantitative improvements translate into clinically more faithful delineation of both fine-grained and morphologically complex targets.

Finally, category-wise evaluation showed that the gains of \methodname{} were especially pronounced for disease-related targets (Fig.~\ref{fig:main_2}e). 
Across lesions, tumors, infarcts, hemorrhages, pneumonia-related abnormalities, pathological cells and skin lesions, \methodname{} consistently achieved higher mean Dice scores than the competing methods. 
This disease-level advantage is particularly important because such targets often exhibit substantial intra-class variation and weak semantic grounding when represented by generic class names alone. 
Together, these results show that \methodname{} is not only broadly effective across modalities and clinical scenarios, but is especially strong on clinically meaningful disease segmentation tasks, motivating a closer examination of the semantic factors underlying this advantage.

\begin{figure}[p]
\centering
\includegraphics[width=\linewidth]{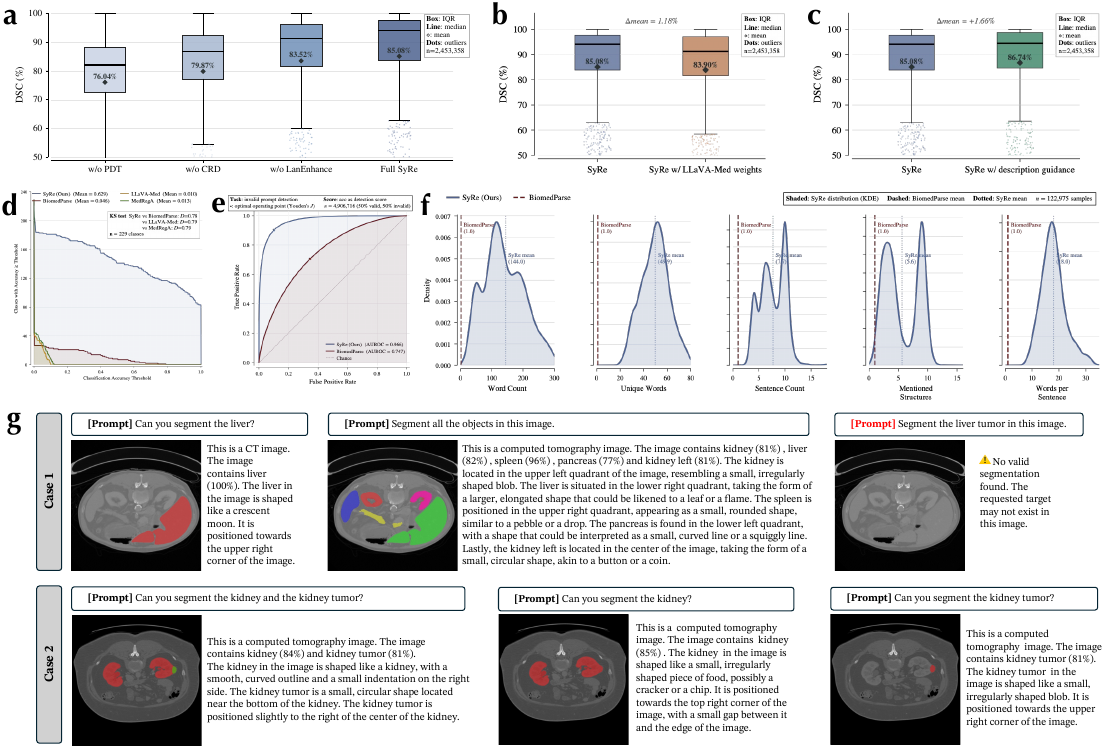}
\caption{
\textbf{Region-level semantic grounding unlocks flexible on-demand analysis through text prompts.}
\textbf{a,} Ablation of PDT, CRD, and LanEnhance on the held-out \textit{SyReData} benchmark.
\textbf{b,} Comparison between the default initialization and initialization with LLaVA-Med weights.
\textbf{c,} Upper-bound analysis of optional description guidance at inference. The additional description in this analysis is derived from the ground-truth segmentation mask and is supplied only to test whether the learned spatial relations can be further exploited when such guidance is available; it is not used in the standard inference protocol or in the other quantitative results.
\textbf{d,}
Reverse cumulative distribution function (1-CDF) showing the number of anatomical classes achieving classification accuracy above varying thresholds.
Each curve represents a different model's performance across the 229 evaluated classes from the internal test set. The area under each count curve, divided by 229, equals the mean per-class classification accuracy. The curves show that SyRe maintains a higher number of classes above the displayed accuracy thresholds than the comparison methods. The inset reports the rechecked two-sample Kolmogorov--Smirnov statistics: $D=0.78$ versus BiomedParse, $D=0.79$ versus LLaVA-Med and $D=0.79$ versus MedRegA ($n=229$ classes per method). 
\textbf{e,} Receiver operating characteristic (ROC) curves evaluating the ability of models to distinguish valid from invalid text prompts. The task involves binary classification of prompts, where valid prompts correspond to correct anatomical or pathological queries (positive class), and invalid prompts represent semantically incorrect or nonsensical queries (negative class). Filled circles indicate optimal operating points determined by Youden's $J$ statistic (maximum of TPR $-$ FPR), representing the threshold that best balances sensitivity and specificity. 
\textbf{f,} Complexity statistics computed per generated description, including word count, unique words, sentence count, mentioned structures and words per sentence. The panel shows the SyRe distribution for $n=122{,}975$ descriptions and the corresponding BiomedParse means.
\textbf{g,} Representative examples. Case 1 shows standard single-target prompting, \textit{seg-all} prompting and invalid-prompt rejection. Case 2 shows both joint prompting (\textit{kidney} and \textit{kidney tumor}) and separate prompting for each target. Numbers in parentheses denote model confidence.
}
\label{fig:main_3}
\end{figure}

\begin{figure}[p]
\centering
\includegraphics[width=0.88\linewidth]{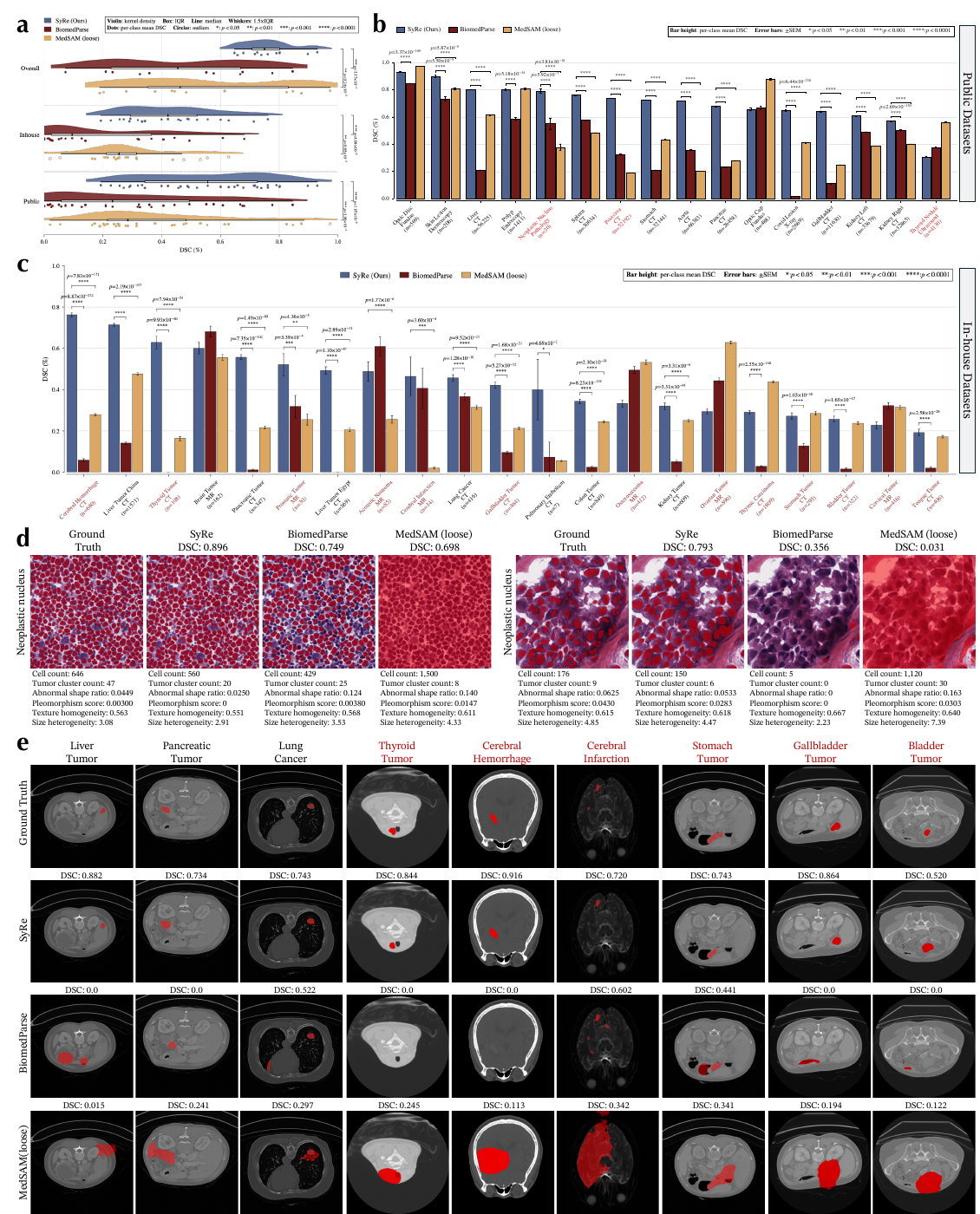}
\caption{
\textbf{\methodname{} maintains accurate disease-target delineation across unseen cohorts, populations and cancer types.}
\textbf{a,} Distribution of per-class Dice scores on the full external benchmark, shown overall and separately for the public and in-house subsets.
\textbf{b,} Per-category comparison for 16 public external targets. Extended Data Fig.~\ref{fig:extended_6} displays the same target categories and additionally shows MedSAM under tight-box, one-point and no-prompt conditions.
\textbf{c,} Per-category comparison on in-house external datasets collected from hospitals in China and Egypt.
\textbf{d,} Two representative neoplastic nucleus cases. In addition to segmentation masks, we applied a published explainable pathology quantification pipeline to the predicted masks to derive downstream pathology descriptors, including cell count, tumor cluster count, abnormal shape ratio, pleomorphism score, texture homogeneity and size heterogeneity. Masks predicted by \methodname{} yield pathology information that is consistently closer to the ground truth than that obtained from BiomedParse or MedSAM (loose).
\textbf{e,} Representative in-house examples across diverse tumor and lesion categories.
}
\label{fig:main_4}
\end{figure}

\subsection{Region-level semantic grounding unlocks flexible on-demand analysis through text prompts}

Given the pronounced gains of \methodname{} on disease-related targets in internal evaluation, we next asked whether these advantages arise from stronger semantic grounding rather than from benchmark-specific optimization alone. 
Across the held-out \textit{SyReData} benchmark, ablation analysis compared the effects of removing PDT, CRD supervision and LanEnhance (Fig.~\ref{fig:main_3}a).

Removing PDT reduced the mean DSC from 85.08\% to 76.04\%; removing CRD supervision reduced it to 79.87\%, and removing LanEnhance reduced it to 83.52\%. The w/o PDT condition also removes description-level supervision because CRD descriptions are generated within the seg-all with CRD format. Its larger drop therefore cannot be attributed to prompt-format diversity alone or used to rank the independent contributions of PDT and CRD. The ablation supports the benefit of the combined prompt-diversified, description-enriched training design, while the separate effects of prompt diversity and description supervision are not fully isolated by these settings.

To further assess whether the observed improvement could simply be inherited from existing medical-domain language pretraining, we replaced the default initialization with LLaVA-Med weights (Fig.~\ref{fig:main_3}b). This modification did not improve performance; instead, the mean DSC decreased from 85.08\% to 83.90\%. A likely explanation is that the language required for our task differs from report-style medical text. The CRD strategy provides simple but spatially grounded descriptions of image content, emphasizing object presence, morphology and relative location, which are directly aligned with segmentation. By contrast, medical-report pretraining emphasizes diagnostic and narrative semantics, which are less well matched to region-level promptable segmentation. In a separate analysis-only upper-bound experiment, adding a description derived from the ground-truth segmentation mask at inference increased mean DSC from 85.08\% to 86.74\% (Fig.~\ref{fig:main_3}c). 
This mask-dependent gain is unavailable under standard mask-free inference; it shows only that the model can exploit additional spatial guidance when a reference mask is available. The ablation and initialization results support the contribution of task-relevant semantic grounding rather than generic medical-domain language priors.

We next evaluated whether \methodname{} retains this semantic advantage when explicit class names are removed. 
In the open-ended \textit{seg-all} setting, where the model is asked to segment all objects in the image without enumerating a named target, \methodname{} preserved substantially stronger performance across classification-accuracy thresholds than competing methods (Fig.~\ref{fig:main_3}d). 
At a per-class classification accuracy threshold of 0.8, 51.09\% (117/229) of the evaluated classes remained above the threshold for SyRe, compared with 0\% (0/229) for BiomedParse, LLaVA-Med and MedRegA (Fig.~\ref{fig:main_3}d). 
This result indicates that \methodname{} is not merely matching class tokens to masks, but is better able to identify and organize semantically meaningful targets under open-ended prompting.

A related requirement for clinically reliable prompting is the ability to reject invalid requests. 
On an invalid-prompt detection task, \methodname{} achieved near-perfect discrimination between valid and invalid prompts and outperformed BiomedParse (Fig.~\ref{fig:main_3}e). MedSAM was not included in this semantic text-prompt comparison because it does not expose an equivalent text-prompt interface. 
This finding is important for practical deployment, because clinically realistic prompting includes not only successful target queries but also ambiguous or absent targets that should be safely rejected rather than hallucinated.

We further quantified the complexity of the descriptions generated by \methodname{} and compared them with those produced by BiomedParse (Fig.~\ref{fig:main_3}f). 

Across the samples shown in Fig.~\ref{fig:main_3}f, \methodname{} consistently generated richer and more structured descriptions, with higher lexical diversity, more mentioned structures, and more complex sentence organization. 
This suggests that \methodname{} has genuinely learned shape, location and inter-structure relationships, rather than relying only on category-level matching. 
These results further support the view that the advantage of \methodname{} arises from stronger region-level semantic grounding rather than from benchmark-specific prompt matching, and they suggest that \methodname{} can also be used to automatically generate informative descriptions for new data, providing a practical route for future large-scale data expansion. Representative examples further illustrate how this semantic advantage translates into on-demand analysis across diverse clinical tasks (Fig.~\ref{fig:main_3}g). 
In Case 1, \methodname{} supports standard single-target segmentation, \textit{seg-all} prompting for joint multi-structure analysis, and reliable rejection of invalid prompts when the requested target is absent. 
In Case 2, \methodname{} supports both joint prompting (\textit{kidney} and \textit{kidney tumor}) and separate prompting for each target, enabling flexible switching between holistic analysis of related structures and focused interrogation of a specific subtarget within the same image. 
These examples show that \methodname{} does not merely improve mask prediction, but supports task-flexible, clinician-driven analysis at different semantic granularities, which is central to on-demand use across diverse clinical tasks. 
If this semantic grounding is indeed the key mechanism underlying the performance gains of \methodname{}, then its benefits should persist under real-world distribution shifts rather than being confined to the internal benchmark. We therefore next asked whether the same advantage extends to unseen datasets, centers, populations and disease categories. 
We further compared SyRe with several representative 3D segmentation models, including SAT, SegVol, and Swin UNETR, on volumetric datasets (Extended Data Fig.~\ref{fig:extended_4}). Despite processing data slice-wise, SyRe still outperformed these 3D baselines. A likely reason is the trade-off between volumetric context and spatial resolution: under GPU memory constraints, 3D models typically operate on low-resolution volumes (for example, $128 \times 128 \times 128$), whereas SyRe can process much higher-resolution 2D slices (up to $1024 \times 1024$). For high-precision segmentation, this higher spatial resolution appears to be more beneficial than explicit 3D context in our current setting.

\begin{figure}[p]
\centering
\includegraphics[width=\linewidth]{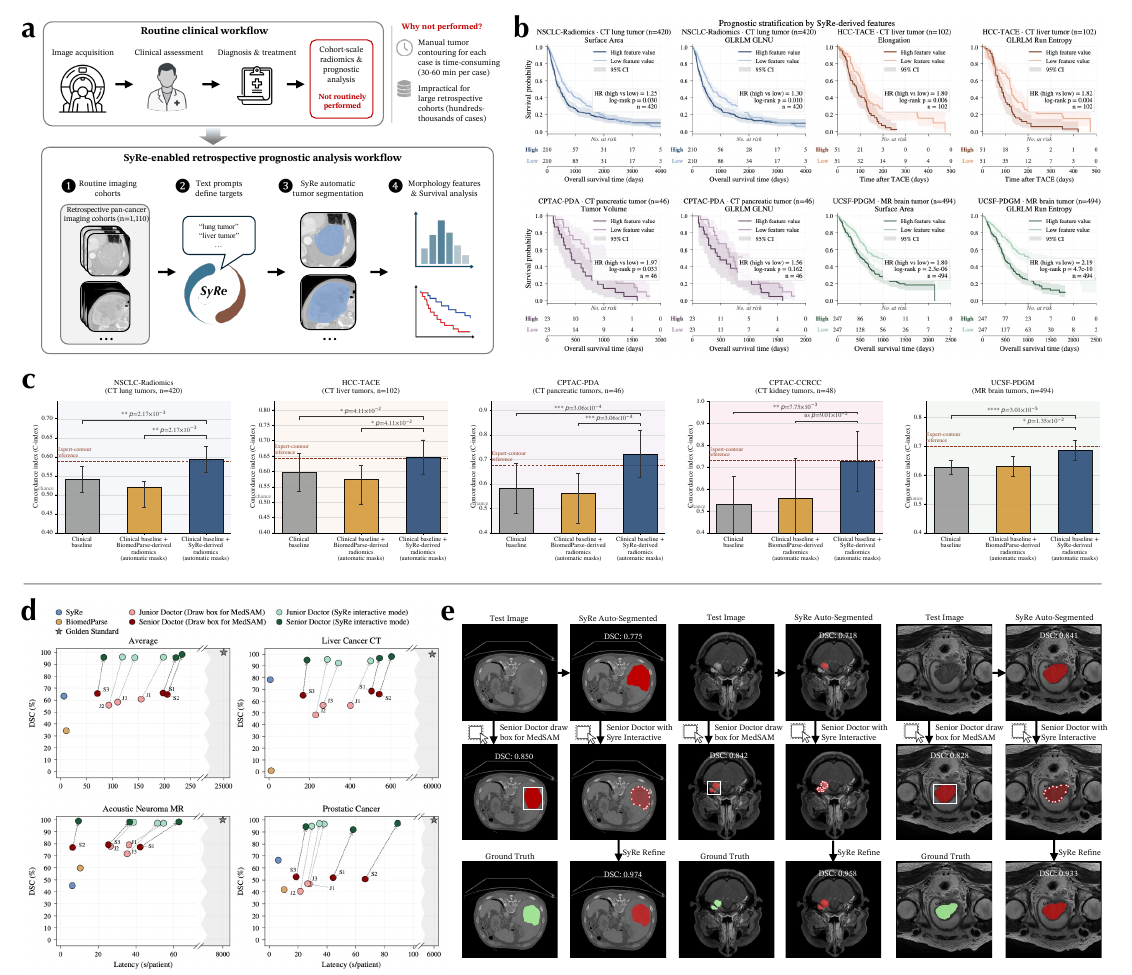}
\caption{
\textbf{\methodname{} enables cohort-scale prognostic analysis from routine oncology images and supports clinician-guided refinement.}
\textbf{a,} Schematic of the \methodname{}-enabled retrospective prognostic-analysis workflow. In routine clinical practice, cohort-scale radiomics and prognostic analysis is not routinely performed because manual tumor contouring for every case is impractical for large retrospective cohorts. \methodname{} uses class-name prompts to generate automatic tumor masks from routine CT and MR cohorts, enabling radiomics extraction and survival analysis without cohort-scale manual contouring.
\textbf{b,} Prognostic stratification using representative \methodname{}-derived morphology and texture features. Kaplan-Meier curves compare survival outcomes between high- and low-feature groups within each cohort.
\textbf{c,} Prognostic modeling using automatically derived radiomics. C-indices are shown for clinical variables alone, clinical variables plus BiomedParse-derived radiomics, and clinical variables plus \methodname{}-derived radiomics. Brackets indicate statistical comparisons. The dashed line denotes the expert-contour reference derived from expert manual contours; it is a practical reference for the cohort and not a theoretical upper bound.
\textbf{d,} Accuracy-latency comparison for automatic segmentation and clinician-in-the-loop refinement across three representative tumor segmentation tasks.
\textbf{e,} Representative clinician-guided refinement examples. Starting from the automatic \methodname{} mask, lightweight refinement produces masks closer to the evaluation reference than MedSAM results based on manually drawn boxes.
}
\label{fig:main_5}
\end{figure}

\subsection{\methodname{} maintains accurate disease-target delineation across unseen cohorts, populations and cancer types}

We next asked whether the semantic advantages of \methodname{} extend beyond internal held-out evaluation to real-world distribution shifts, including unseen datasets, centers, patient populations and disease categories. 

The distributions of per-class Dice scores indicate an overall advantage for \methodname{} in the displayed public and in-house comparisons (Fig.~\ref{fig:main_4}a), but the category-level results show that this advantage is not uniform across targets (Fig.~\ref{fig:main_4}b,c).

On the 16 public external targets displayed in Fig.~\ref{fig:main_4}b, \methodname{} achieved higher mean DSC than MedSAM with loose boxes on COVID lesions, neoplastic nuclei, skin lesions and most of the displayed abdominal targets, but lower mean DSC on optic cup, optic disc, polyp and thyroid nodule. Extended Data Fig.~\ref{fig:extended_6} displays the same 16 categories and sample counts. The tight-box, one-point and no-prompt comparisons show that the relative performance also depends on the spatial information supplied to MedSAM: MedSAM benefits from more precise spatial prompts on several targets. These results should therefore be interpreted with the prompting conditions specified, and do not establish a uniform advantage across public targets.

The multinational in-house datasets collected from hospitals in China and Egypt also showed heterogeneous performance (Fig.~\ref{fig:main_4}c). \methodname{} outperformed both displayed baselines on several targets, including cerebral infarction, liver tumor, thyroid tumor and pancreatic tumor. However, ovarian tumor, thymic carcinoma, bladder tumor, cervical tumor and tongue tumor remained difficult. MedSAM with loose boxes achieved higher mean DSC on ovarian tumor and thymic carcinoma, while both displayed baselines exceeded \methodname{} on cervical tumor. These results support transfer across some clinical settings while identifying target-specific limitations.

To further assess whether these segmentation gains preserve clinically relevant downstream information, we analyzed two representative neoplastic nucleus cases using a published explainable pathology quantification pipeline that derives downstream morphologic descriptors from tissue and nuclei segmentation masks~\cite{liu2024panoptic} (Fig.~\ref{fig:main_4}d). In both cases, \methodname{} achieved the highest segmentation accuracy and, more importantly, produced pathology measurements that were consistently closer to the ground truth than those obtained from BiomedParse or MedSAM. These derived descriptors included cell count, tumor cluster count, abnormal shape ratio, pleomorphism score, texture homogeneity and size heterogeneity. Thus, the advantage of \methodname{} in pathology is not limited to improved visual overlap, but extends to more faithful recovery of diagnostically relevant quantitative information from H\&E images. This is particularly important because clinically useful pathology segmentation must preserve the morphologic and architectural cues that underpin downstream interpretation.

Representative in-house examples across diverse tumor and lesion categories further illustrate these advantages (Fig.~\ref{fig:main_4}e). 

For the displayed examples, \methodname{} obtained DSC values of 0.882 for liver tumor, 0.734 for pancreatic tumor, 0.743 for lung cancer, 0.844 for thyroid tumor, 0.916 for cerebral hemorrhage, 0.720 for cerebral infarction, 0.743 for stomach tumor, 0.864 for gallbladder tumor and 0.520 for bladder tumor. These illustrative cases are not cohort-level estimates; the category-level comparisons show that external segmentation accuracy remains dependent on the target and evaluation setting.

\subsection{A single text-driven model enables cohort-scale prognostic analysis from routine oncology images}

Having established that \methodname{} generalizes across unseen datasets, populations and tumor types, we next asked whether automatic text-prompted tumor segmentation could support a clinically meaningful task that is rarely performed in routine clinical workflow: retrospective quantitative oncology analysis at cohort scale. Large retrospective imaging cohorts are often linked to survival, recurrence or treatment-response outcomes, but extracting tumor morphology and radiomics features from such cohorts traditionally requires manual delineation of every lesion. This requirement makes cohort-scale radiomics analysis impractical in many routine clinical and retrospective research settings. We therefore evaluated whether a single \methodname{} model, using simple class-name prompts without task-specific retraining, could convert routine CT and MR tumor imaging cohorts into quantitative radiomics datasets for prognostic analysis (Fig.~\ref{fig:main_5}a).

We first evaluated whether \methodname{}-derived radiomics retain prognostic information across five retrospective tumor imaging cohorts spanning CT and MR ($n=1{,}110$). Representative morphology and texture features derived from \methodname{}-generated masks stratified survival outcomes across NSCLC-Radiomics, HCC-TACE, CPTAC-PDA, and UCSF-PDGM (Fig.~\ref{fig:main_5}b). These results show that automatically generated tumor masks preserve clinically relevant imaging phenotypes across heterogeneous cancer types and imaging modalities, including lung, liver, pancreatic, kidney and brain tumors.

We further tested whether \methodname{}-derived radiomics add prognostic value beyond routinely available clinical variables. For each cohort, we compared models using clinical variables alone, clinical variables plus BiomedParse-derived radiomics from automatic masks, and clinical variables plus \methodname{}-derived radiomics from automatic masks. Across all five cohorts, adding \methodname{}-derived radiomics improved prognostic discrimination over the clinical baseline, with relative C-index gains of 8.5-37.2\% across cohorts (all adjusted $P<0.05$), and \methodname{}-derived radiomics consistently outperformed the corresponding BiomedParse-derived radiomics model (Fig.~\ref{fig:main_5}c).

These results define the clinical value of \methodname{} more precisely. The key advantage is not that \methodname{} replaces expert manual delineation for individual cases, but that it enables a clinically meaningful analysis that is otherwise difficult to perform at cohort scale. By using a single text-driven model without tumor-specific retraining or expert-provided spatial initialization, \methodname{} can extract prognostically informative radiomics from routine oncology images and support retrospective survival analysis without requiring manual tumor contouring for every case. Thus, \methodname{} provides a scalable route for quantitative oncology analysis in settings where cohort-scale manual delineation is impractical.

\subsection{Lightweight clinician-guided refinement improves tumor segmentation beyond box-based interaction}

Whereas retrospective cohort analysis emphasizes scalability, individual clinical cases may require higher-precision masks than those produced by a single automatic pass. We therefore evaluated whether \methodname{} can support clinician-in-the-loop refinement when users need to correct or improve an initial segmentation. This setting tests a complementary form of clinical utility: efficient case-level correction rather than cohort-scale automation.

Across the average setting and three representative tumor segmentation tasks - liver cancer CT, acoustic neuroma MR and prostate cancer - \methodname{} provided a strong automatic segmentation baseline with minimal latency (Fig.~\ref{fig:main_5}d). Lightweight refinement by junior or senior doctors further improved segmentation accuracy and consistently outperformed box-based interaction with MedSAM. These gains were achieved with only a modest increase in latency, indicating that \methodname{} can reduce the burden of manual spatial prompting while preserving an explicit route for clinician correction when the initial automatic mask is imperfect.

Representative examples further illustrate this workflow advantage (Fig.~\ref{fig:main_5}e; Extended Data Fig.~\ref{fig:extended_10}). In a liver cancer CT case, the automatic \methodname{} mask achieved a DSC of 0.775 and improved to 0.974 after clinician-guided refinement, compared with 0.850 for senior-doctor box prompting with MedSAM. In an acoustic neuroma MR case, refinement improved the DSC from 0.718 to 0.958, compared with 0.842 for MedSAM-based interaction. In a prostate cancer case, refinement improved the DSC from 0.841 to 0.933, compared with 0.828 for MedSAM-based interaction.

Together, these results show that \methodname{} is not a one-shot segmentation system whose output must simply be accepted or rejected. Instead, it supports two complementary modes of clinical use: cohort-scale automatic analysis for retrospective quantitative oncology studies, and efficient clinician-guided correction when individual cases require higher segmentation precision.

\section{Discussion}

A central challenge in medical image analysis is that language supervision is often too weak to support fine-grained spatial understanding. 
In existing language-driven systems, text is frequently reduced to a generic class label, which may be sufficient for coarse recognition but is poorly matched to the heterogeneity of clinically meaningful targets. 
This limitation is especially severe for disease-related findings, whose morphology, size, texture, and location can vary substantially even within the same category. 
As a result, prior approaches often struggle to translate text prompts into accurate delineation of anatomical structures and lesions, limiting their usefulness in clinical workflows that require both semantic precision and spatial fidelity.

Our study suggests that this bottleneck can be addressed by strengthening region-level semantic grounding. 
Rather than treating language as a shallow prompt attached to a segmentation model, \methodname{} uses synergistic vision-language reinforcement to couple semantic understanding with spatial prediction, and further uses the CRD strategy to provide richer supervision about object identity, morphology, and relative position. 
A second component is how language is presented during training: diversified prompt forms expose \methodname{} to different requests rather than a single fixed query per image, encouraging it to condition on what is actually being asked rather than on what the image happens to contain.

In our ablation, the w/o PDT condition produced the largest observed degradation in segmentation accuracy, but it also removed description-level supervision because CRD descriptions are generated within the seg-all with CRD format. This result therefore supports the benefit of the combined prompt-diversified and description-enriched training design, but does not isolate the independent effect of prompt diversity or establish a strict ranking of component contributions.

Together, these components shift language-driven segmentation from category-level matching towards semantically grounded image understanding. 
The resulting model improved segmentation performance across diverse benchmarks, with particularly strong gains on disease-related targets, where weak semantic grounding is most problematic. 
These findings support the view that the major barrier in promptable medical segmentation is not merely model scale, but the mismatch between generic labels and the fine-grained semantic information required for clinically relevant delineation.

This training design also changes what the model can be asked to do. 
Beyond segmentation accuracy, \methodname{} provides a more flexible interface for on-demand medical image analysis. 
It supports single-target prompting, multi-structure prompting, rejection of invalid prompts, and both joint and separate analysis of related targets within the same image. 
This flexibility is important because real clinical use rarely follows a single fixed task template. 
Clinicians may need a broad anatomical overview, focused lesion interrogation, cohort-level feature extraction, or refinement of an initially acceptable result. 
In this sense, \methodname{} functions not only as a stronger segmentation model, but as a task-flexible interface that can convert diverse clinical intents into spatial outputs.

Another important finding is that this semantic advantage translated into robust external generalization. 
\methodname{} maintained competitive performance across unseen public and in-house datasets, multiple cancer and lesion categories, and cohorts from different centers and populations. 
These results suggest that semantically enriched supervision does not merely improve in-distribution fitting, but helps the model learn representations that are more transferable under real-world distribution shifts. 
The pathology examples further show that improved masks can yield more plausible downstream quantitative descriptors, indicating that the benefit of improved segmentation may extend beyond overlap metrics to morphology-aware analysis.

A key clinical implication is that the value of promptable segmentation should not be judged by spatial overlap alone. 
In many clinical research settings, segmentation is an enabling step rather than the final endpoint: clinicians need quantitative measurements of tumor burden, morphology, texture, and spatial extent that can be linked to survival, recurrence, or treatment response. 
Manual delineation remains the reference standard for high-stakes patient-specific decisions, and clinicians can perform it for selected cases. 
However, contouring large retrospective cohorts solely for morphologic or prognostic analysis can require hundreds of hours of expert time. 
Across five retrospective tumor imaging cohorts spanning CT and MR ($n=1{,}110$), \methodname{}-generated masks preserved clinically informative quantitative features that stratified survival, while radiomics derived from these masks improved prognostic modeling over clinical baselines and consistently outperformed the corresponding BiomedParse-derived radiomics models. 
Thus, \methodname{} enables retrospective quantitative oncology analyses that are feasible in principle but impractical to perform manually at the cohort scale. These retrospective results establish feasibility of extracting prognostically informative radiomics from automatically generated masks; they do not demonstrate changes in treatment decisions, patient outcomes or prospective clinical practice. These results are limited to retrospective cohort-level analysis.

Our results also point to a practical path for scaling language-driven medical segmentation. 
The generated descriptions used in \methodname{} are intentionally simple and spatially grounded, focusing on what structures are present, how they are shaped, and where they are located. 
This form of supervision is well matched to segmentation and differs from report-style medical language, which is often richer diagnostically but less directly aligned with region-level delineation. 
Because \methodname{} can itself generate structured descriptions for new data, the framework suggests a scalable pipeline in which semantically enriched text supervision can be iteratively expanded beyond the currently annotated corpus.

Several limitations should be noted. 
First, 
comparisons with fixed-vocabulary baselines are partly affected by vocabulary coverage. BiomedParse supports a predefined class vocabulary, so its performance on the full SyReData benchmark reflects both segmentation quality on shared targets and the inability to address targets outside that vocabulary. Accordingly, the shared-vocabulary subset should be read as the more direct capability comparison, whereas the full-benchmark result measures practical end-to-end coverage and segmentation. 
Second, external performance remains heterogeneous. Ovarian tumor, thymic carcinoma, bladder tumor, cervical tumor and tongue tumor remained difficult in the in-house evaluation, and the earlier public thyroid-nodule comparison did not show an advantage for \methodname{}. Relative performance also depends on the spatial information supplied to the comparator, with MedSAM benefiting from tight boxes on several public targets. Thus, the aggregate comparisons do not establish uniform performance across diseases. 
The split procedures establish held-out membership relative to the SyRe training manifests, with patient- or scan-level separation where within-scan continuity applies, but they do not establish complete source-level independence. Some public benchmarks may originate from source collections related to datasets used elsewhere in the medical-imaging ecosystem, and the provenance of data used to pretrain foundation components cannot be completely audited. We therefore cannot exclude source-level overlap or prior exposure through pretrained components; the external benchmark should be interpreted as held out with respect to the SyRe training manifests rather than as a guarantee of complete source-level independence.\ Third, handcrafted radiomic features vary in their sensitivity to contour quality, imaging protocols, and preprocessing choices; therefore, our results should not be interpreted as showing that every radiomic descriptor can be perfectly recovered from automatic masks. 
Rather, \methodname{} preserved key tumor burden, morphology, and texture signals that were sufficient for cohort-level prognostic analysis. 
Fourth, the prompt forms used during training, although more diverse than a single fixed query, represent a designed distribution of clinical requests. 
Finally, the inference-time confidence filter was selected empirically and has not yet been prospectively calibrated or tested in a clinical deployment setting.

In summary, our work shows that semantically grounded, text-driven segmentation can support clinically useful medical image analysis beyond static mask generation. 
By combining synergistic vision-language reinforcement, scalable region-level semantic supervision and prompt-diversified training, \methodname{} enables accurate promptable segmentation, robust external generalization, multi-cancer retrospective prognostic analysis from automatic tumor masks, and efficient clinician-guided refinement. 
More broadly, these findings suggest that medical foundation models should not only improve benchmark performance but also convert diverse clinical intents into spatial and quantitative outputs that can support real-world research and clinical workflows.

\section{Methodologies} \label{sec:method}

\subsection{Dataset Curation} 

\label{sec:data}

\noindent\textbf{Color Region Description Annotating Strategy.}
A large number of medical image segmentation datasets exist with image-mask pairs. However, datasets for language-driven segmentation tasks are scarce. The most straightforward way to build this kind of dataset is using the class names to construct the image-mask-label triplets, like BiomedParse did~\cite{natFM_zhao2024foundation}.
However, this straightforward strategy does not establish a robust relationship among the image, mask, and category names, leading to inferior results, as shown by the \textit{w/o CRD} ablation in Fig.~\ref{fig:main_3}a and by the external comparison in Fig.~\ref{fig:main_4}a.
The issue arises from a gap between the semantic information conveyed by the category name and the morphological information represented by the masks. 
Category names often fail to provide details about the location and shape of anatomical structures, which are crucial for a comprehensive understanding of their morphology. 
For example, the pancreas undergoes significant shape and location changes across different CT slices; aligning all these variations to a single term, \textit{pancreas}, is quite challenging for the foundation model.
Thus, we want to obtain the slice-wise description to better guide the model. 
For each image and mask pair, we leverage the InternVL-1.5 to generate their corresponding text descriptions. Generally, the off-the-shelf large vision-language models (VLMs) such as GPT4, InternVL, QWenVL cannot understand the medical images, i.e., the generated text descriptions are weird. However, they are powerful enough to handle very simple tasks, such as describing different color patches. 
As illustrated in Fig.~\ref{fig:main_1}d, the CRD strategy involves taking 2D masks as inputs and converting each category into distinct pre-defined colors. We then input these colored masks into the VLFMs to generate diverse and satisfactory descriptions of the shapes and relative positions of all the colored regions.
We showcase several generated descriptions in Extended Data Fig.~\ref{fig:extended_1}. Using the CRD strategy, we converted the eligible image-mask pairs into approximately 20 million image-mask-description triplets spanning nine imaging modalities and 229 segmentation targets.

\noindent\textbf{Collecting Multi-disease Data from Hospitals.} 
Public datasets mainly target organ segmentation; there are few disease segmentation datasets, especially for cancer.
Thus, as a supplement to the public datasets, we collected a comprehensive disease dataset. 
The dataset was collected from Sun Yat-sen Memorial Hospital, Sun Yat-sen University, Guangdong, China. 
The dataset consists of 20 common human diseases, especially cancers, covering all the human body systems.
The datasets used in the previous text-driven models lack this disease, making them impractical for clinical use.
Thus, we built an in-house external test set to test the potential clinical usage of the proposed model. This test set contains 20 target categories, comprising 17 cancer types and 3 non-cancer diseases (cerebral infarction, cerebral haemorrhage and pulmonary embolism),\ covering all the systems in the human body: Central Nervous System (brain tumor, cerebral infarction, cerebral hemorrhage), Head \& Neck (acoustic neuroma, tongue cancer, thyroid cancer), Respiratory System (lung cancer, pulmonary embolism), Circulatory System (thymic carcinoma), Digestive System (stomach cancer, pancreas cancer, gallbladder cancer, liver cancer, colon cancer), Urinary and Reproductive System (bladder cancer, prostate cancer, kidney cancer, ovarian cancer, cervical cancer), Musculoskeletal System (osteosarcoma), each disease contains 20 patients from Sun Yat-sen Memorial Hospital, Sun Yat-sen University. 
For cross-model evaluation on this cohort, the text-prompt methods SyRe and BiomedParse used target names normalised before inference with a fixed mapping, including case and underscore normalisation and the cancer-to-tumor substitutions required by the SyRe vocabulary. MedSAM was evaluated with the corresponding target-specific spatial prompt (the prescribed box or point for that category), rather than with a text prompt. The mapping changed only the text prompt string for the text-prompt methods and did not alter images, reference masks or predicted masks.
The collected MR and CT images come from a variety of imaging devices. The MR images are captured using machines from Philips (Ingenia 1.5T, Ingenia 3.0T, Achieva 3.0T, Ambition 1.5T) and Siemens (MAGNETOM Skyra 3.0T, MAGNETOM Vida 3.0T, MAGNETOM Avanto 1.5T). The CT images are obtained from Siemens (SOMATOM Force, SOMATOM Sensation 64), United Imaging (uCT780), and GE (Discovery HD, Revolution EVO).

\subsection{\methodname{}: Network Architecture}

\methodname{} comprises four modules: a large language model, an image encoder, a text prompt encoder and a mask decoder. To couple high-level semantic reasoning with fine-grained spatial prediction, we instantiate the language backbone $\mathcal{L}$ with the 7B-parameter Vicuna model~\cite{vicuna2023}, which provides a favorable balance between representational capacity and computational efficiency. For visual encoding, rather than adopting a CLIP-style encoder~\cite{radford2021clip}, we use the SAM image encoder~\cite{kirillov2023sam}, denoted by $\mathcal{V}$, because its higher input resolution and stronger pixel-level representation are better suited to dense medical segmentation. We initialize $\mathcal{V}$ from the pre-trained SAM encoder, and build both the prompt encoder and the mask decoder on a SAM-style decoding architecture.

\noindent\textbf{Synergistic vision-language reinforcement (LanEnhance).}
Given an input image $\mathbf{I} \in \mathbb{R}^{H \times W \times 3}$, the image encoder produces a spatial feature map
$
\mathbf{E}_v = f_{\mathrm{enc}}(\mathbf{I}) \in \mathbb{R}^{C \times h \times w},
$
where $C=256$ and $h=w=64$. In addition to the spatial embedding used for mask prediction, the final-layer visual tokens $\mathbf{E}_v^{\mathrm{last}} \in \mathbb{R}^{B \times L \times d_v}$ are projected into the language space through a vision-to-language projection layer $p_{v\rightarrow l}$ and concatenated with the text prompt tokens. This forms the vision-to-language pathway, allowing the large language model to condition generation on image content. For an image $x_i$ and a text instruction $x_l$, the projected visual representation is integrated by the language model to produce the textual output
$y_l = \mathcal{L}\bigl(p_{v\rightarrow l}(E_v), x_l\bigr),$
thereby aligning visual evidence with linguistic semantics.

To translate this semantic grounding back into spatial prediction, we introduce a complementary language-to-vision pathway. Specifically, the projected visual-language representation is mapped back into the decoder feature space with a language-to-vision projection layer $p_{l\rightarrow v}$, yielding
$
E_{l\rightarrow v} = p_{l\rightarrow v}(E_{v\rightarrow l}),
$
which is then fused with the original visual embedding $E_v$ before decoding. This bidirectional interaction reinforces correspondence between image regions and linguistic descriptions, enabling the decoder to localize targets more precisely under free-text instructions. We refer to this bidirectional design as the LanEnhance module, and ablate it in Fig.~\ref{fig:main_3}a.

To explicitly couple language generation with segmentation, we augment the vocabulary with a special token \texttt{[SEG]}. During training and inference, prompts such as \texttt{The <image> provides an overview of the image. Can you segment the \{class name\} in this image?} encourage the model to generate responses containing \texttt{[SEG]} tokens associated with the queried anatomical or pathological entities. Here, \texttt{<image>} is replaced by the visual tokens derived from the SAM encoder, and \texttt{\{class name\}} denotes the target specified by the user. The hidden states corresponding to \texttt{[SEG]} tokens, denoted $E_{\mathrm{seg}}$, are projected to the mask-decoder space and used together with the reinforced visual features to predict binary masks:
\begin{equation}
y_v = \mathcal{M}\bigl(p_{v\rightarrow l}(E_{\mathrm{seg}}),\, E_v + E_{l\rightarrow v}\bigr),
\qquad \{y_v\}_i \in \{0,1\}.
\end{equation}
Through this end-to-end design, \methodname{} establishes a tighter coupling between image content and language supervision, allowing it to generate segmentation outputs that are semantically consistent with the user instruction.

\noindent\textbf{\methodname{} interactive inference and multi-round mask auto-refinement.}
We implemented \methodname{} as an interactive application that takes a clinical image and a free-text instruction as input, performs joint language generation and grounded segmentation in a single forward pass, and displays the predicted masks as semi-transparent overlays on the original image. Each mask is assigned a distinct colour and is paired with a confidence score shown alongside the corresponding anatomical label in the textual response. When the confidence mechanism described above indicates that all predicted masks for a given prompt fall below the acceptance threshold, the system suppresses the segmentation output and instead returns a natural-language warning, thereby avoiding the display of anatomically implausible results. To support real-time refinement, we cache the full-resolution logit map $\mathbf{L} \in \mathbb{R}^{H \times W}$ produced by the SAM mask decoder after each inference call, store it in memory in 16-bit floating-point format, and transmit it to the client together with the binary mask. The displayed mask is obtained by zero-thresholding, $\mathcal{M} = \{(i,j): L_{ij} > 0\}$. Because this logit map preserves the decoder's dense spatial confidence landscape rather than only a hard decision boundary, it provides a natural substrate for post hoc correction without requiring an additional model forward pass, which would otherwise be too slow for practical interactive use. The interface therefore includes a control-point-based editor that allows users to refine any predicted mask by dragging, adding, or deleting control points along its boundary. The edited result $\mathcal{M}^{\prime}$ is returned to the server as a binary image, with foreground pixels satisfying $M^{\prime}_{ij} > 128$. To convert these discrete edits into a continuous correction signal in logit space, we construct an asymmetric soft signed-distance field (SDF) from $\mathcal{M}^{\prime}$. Specifically, let $d^{+}_{ij}$ denote the Euclidean distance from pixel $(i,j)$ to the nearest background pixel when $(i,j) \in \mathcal{M}^{\prime}$, and let $d^{-}_{ij}$ denote the Euclidean distance to the nearest foreground pixel when $(i,j) \notin \mathcal{M}^{\prime}$; both maps are computed using the exact Euclidean distance transform.

\noindent Let $q=\max(H,W)$, $\tau_{+}=0.02q$ and $\tau_{-}=0.01q$. The implementation computes
\[
s_{ij}=\begin{cases}
\{1+\exp[-d^{+}_{ij}/\tau_{+}]\}^{-1}-\tfrac12,&M^{\prime}_{ij}>128,\\
\{1+\exp[d^{-}_{ij}/\tau_{-}]\}^{-1}-\tfrac12,&M^{\prime}_{ij}\le128,
\end{cases}
\qquad
\mathbf{L}_{\mathrm{ref}}=\mathbf{L}+120\,\operatorname{Resize}_{\mathrm{bilinear}}(\mathbf{s};W_L,H_L),
\]
where $\mathbf{L}$ is the cached decoder logit (decoded from float16 to float32 for this computation), and $(H_L,W_L)$ is its spatial size. The final mask is $\mathcal{M}_{\mathrm{ref}}=\{(i,j):L_{\mathrm{ref},ij}>0\}$. The SDF correction is applied once per refinement request to the cached original logit map; a subsequent request therefore represents a new edit of that cached map rather than an iterative update of the logit itself.

\subsection{Training Protocol and Experimental Setting}

During data pre-processing, we obtained approximately 20M medical image-mask-text triplets for model development and validation. For the two-dimensional run reported here, the training and held-out manifests are provided in the linked public dataset release.

For internal validation, we split data at the patient or scan level where within-scan continuity exists, including CT and MRI series and consecutive frames, to prevent slices from the same scan appearing across training and held-out evaluation. For external validation, all datasets were held out and did not appear during model training. These datasets provide a stringent test of generalization to new patients, imaging conditions and segmentation tasks.

\noindent\textbf{Prompt-diversified training (PDT).}
Training a language-driven segmentation model with a single instruction format couples the model to one response template and creates a mismatch with the shorter, more varied prompts encountered at inference. We therefore train \methodname{} with three interleaved instruction formats. In the \textit{seg-all with CRD} format, the model is asked to describe the image and segment every annotated target, with the corresponding CRD descriptions included in the assistant response. In the \textit{seg-all} format, the request lists all annotated target names and the assistant returns their segmentation masks without CRD descriptions. In the \textit{seg-subset} format, a subset of the annotated targets is sampled and those target names are listed in the request. The three formats are sampled with probabilities $0.4$, $0.4$, $0.2$ at each training step; an audit of the training manifest against the CRD description map found coverage of 100\% (3,094,486/3,094,486 entries), so these are the global training-sample proportions. In the \textit{seg-subset} format the subset size $k$ is drawn uniformly from $\{1,\dots,N\}$, where $N$ is the number of annotated targets present in the slice, after which $k$ targets are sampled without replacement. The public training loader now implements this policy explicitly; held-out evaluation bypasses the training sampler and uses class-name prompts without CRD descriptions.

A single-target request is therefore the special case $k=1$ within \textit{seg-subset}, rather than a fourth training format. The testing protocol is separate from this training sampler: standard quantitative evaluation uses a comma-separated prompt containing all target names, while the released inference interface additionally supports single-target, multi-target and open-ended \textit{seg-all} requests. 

Retaining the multi-target formats is deliberate. CRD descriptions are relational, in that they locate each region with respect to the other structures present in the same slice, so decomposing every image into independent single-target samples would remove the co-occurring context that makes such descriptions meaningful. Multi-target training also exposes the model to the joint queries it must serve at inference, and anatomical co-occurrence constrains the set of plausible predictions, which supports the rejection of targets that are absent from the image.

Unless otherwise stated, quantitative evaluation uses a single prompt in which the names of all target classes are concatenated with commas, for example \texttt{Can you segment the \{class name 1\}, \{class name 2\}, \{class name 3\}?}, corresponding to the multi-target setting illustrated in Fig.~\ref{fig:main_3}g. No description guidance is supplied in this standard protocol or in the other quantitative results. Benchmark metrics are computed from the raw predicted masks before applying the inference-time confidence filter; the filter is an interface safeguard and is not used to selectively remove cases from the reported benchmark scores. The released inference interface also supports open-ended \textit{seg-all} requests, in which the user asks the model to segment all objects without enumerating class names. Figure~\ref{fig:main_3}c is a separate upper-bound analysis in which an additional description derived from the ground-truth segmentation mask is supplied only to test whether the spatial relations learned during training can be exploited when such guidance is available; this mask-derived condition is not available in deployment and is not required for the standard protocol.

\noindent\textbf{Ablation settings.}
To isolate the contribution of each component, we retrained \methodname{} under three ablated configurations and evaluated all models on the held-out \textit{SyReData} benchmark ($n = 2{,}453{,}358$) using an identical inference protocol (Fig.~\ref{fig:main_3}a).
In the \textit{w/o PDT} setting, the three interleaved instruction formats were replaced by a single fixed single-target format, so that each training sample requested exactly one named target per image.
In the \textit{w/o CRD} setting, all CRD descriptions were removed from the assistant responses and the model was supervised with class names alone, while the prompt-format sampling was left unchanged.
In the \textit{w/o LanEnhance} setting, the language-to-vision pathway $p_{l\rightarrow v}$ was removed and the mask decoder operated on the visual embedding $E_v$ alone, retaining the vision-to-language pathway used for text generation.
All other architectural choices, data splits, optimisation settings and training budgets were held identical across the four configurations.

We note that the \textit{w/o PDT} and \textit{w/o CRD} settings are not fully orthogonal: because CRD descriptions are generated within the \textit{seg-all with CRD} format, restricting training to a single-target format also reduces the amount of description-level supervision seen during training. The \textit{w/o PDT} result should therefore be interpreted as the combined effect of restricting the model to a single query form, rather than as an effect of prompt format in isolation.

\subsection{Implementation Details}

The experiments used 40 distributed processes, a per-process micro-batch size of 8 and 10 gradient-accumulation steps, giving a global training batch size of 3,200. We used NVIDIA H800 GPUs, BF16 precision and DeepSpeed ZeRO stage 2. Our vision-language framework is inspired by GLaMM~\cite{rasheed2024glamm}, utilizing 2-layer MLPs with GELU activation for the V-L and L-V projection layers, similar to LLaVA-v1.5~\cite{liu2023llava}. We initialize the vision modules using SAM with ViT-H weights~\cite{kirillov2023sam}.
The reported run used AdamW with an initial learning rate of $1\times10^{-4}$, weight decay 0.01 and betas $(0.9,0.999)$. The learning rate used a linear warm-up for 100 optimization steps to a maximum of $1\times10^{-4}$. The scheduler was configured with a 5,000-step decay horizon ($\texttt{total\_num\_steps}=5000$), which is a scheduler parameter rather than the total duration of the run. The continuation log records optimization steps through step 7,740, with the learning rate at zero after the scheduler horizon; the final saved checkpoint documented for the reported run is $\texttt{global\_step7728}$. The loss combined autoregressive cross-entropy, per-pixel binary cross-entropy, Dice and boundary terms. The LLM was fine-tuned with LoRA rank 16; the remaining trainable vision, projection and mask-decoder modules followed the released training configuration.
The text instructions are formulated in a predefined conversation format. For the single-target case:
``\texttt{A chat between a curious human and an artificial intelligence assistant. \ldots USER: Can you segment the \{class name\} in this \{modality\} image? ASSISTANT: This is a <p> \{modality\} </p> image. The image contains <p> \{label\} </p> [SEG].}''
For the \textit{seg-all} case, the same class-name template is used with the comma-separated names of all annotated targets. For the \textit{seg-subset} case, it is used with the sampled subset of target names. For the \textit{seg-all with CRD} case, the request asks for a detailed image description with interleaved segmentation masks and the assistant response additionally contains the CRD description of each segmented region, in the form
``\texttt{ASSISTANT: This is a <p> \{modality\} </p> image. The image contains <p> \{label 1\} </p> [SEG] \ldots \{CRD description\}}''.
We add the token \texttt{[SEG]} for the segmentation task, which is a 1D token further processed by the prompt encoder of SAM; one \texttt{[SEG]} token is emitted per requested target.
We use the Dice similarity coefficient (DSC, \%) as the primary evaluation metric, calculated using the definitions of true positive (TP), false positive (FP), and false negative (FN), given by $\text{DSC}(\hat{V}, V) = \frac{2\text{TP}}{2\text{TP} + \text{FP} + \text{FN}}$.

\subsection{Inference-time mask confidence scoring}

To suppress spurious segmentation outputs arising from out-of-distribution or anatomically inconsistent prompts, we introduce a lightweight confidence scoring mechanism that operates on the full-resolution logit map produced by the SAM mask decoder, requiring no additional model forward pass. For each segmentation token predicted by the language model, the mask decoder outputs a logit map $\mathbf{L} \in \mathbb{R}^{H \times W}$, from which a binary mask $\mathcal{M} = \{(i,j): L_{ij} > 0\}$ is derived by zero-thresholding. Masks whose foreground area ratio $r = |\mathcal{M}|/(HW)$ falls below $0.1\%$ or exceeds $90\%$ of the image area are immediately rejected as degenerate, as such masks typically correspond to either empty predictions or near-complete image flooding, both of which are characteristic of decoder failure modes. For the remaining masks, we characterize the distribution of logit activations within the foreground region by computing the mean $\mu$ and standard deviation $\sigma$ over $\{L_{ij}: (i,j) \in \mathcal{M}\}$. Two complementary scores are then derived: an activation strength score $s_{\text{mean}} = \tanh(\mu / 6)$, which rewards high absolute logit values inside the mask and reflects the decoder's overall certainty about the foreground region, and a spatial consistency score $s_{\text{SNR}} = \sigma(2(\mu/\sigma - 3.5))$, where $\sigma(\cdot)$ denotes the sigmoid function, which rewards a high signal-to-noise ratio and penalizes masks whose foreground activations are spatially heterogeneous-a pattern commonly observed when the decoder is forced to segment a target absent from the image. The composite confidence score is defined as a weighted combination of these two signals, $c = 0.85\, s_{\text{mean}} + 0.15\, s_{\text{SNR}}$, with the higher weight assigned to activation strength as the primary discriminator. Masks with $c < 0.40$ are discarded from the final output; this threshold was determined empirically by inspecting the score distributions of confirmed valid and invalid segmentations across representative cases spanning multiple imaging modalities. When all masks associated with a given prompt are discarded, the system returns a natural-language notification to the user indicating that no confident segmentation target was identified, thereby preventing the display of anatomically implausible results.

\subsection{Retrospective multi-cancer quantitative imaging and prognostic analysis}

To evaluate whether \methodname{}-generated masks preserve clinically meaningful quantitative information and support downstream prognostic analysis at cohort scale, we performed retrospective imaging-feature analysis across five tumor imaging cohorts spanning CT and MR, comprising a total of 1,110 patients: NSCLC-Radiomics for CT lung tumors ($n=420$)~\cite{NSCLC}, HCC-TACE for CT liver tumors ($n=102$)~\cite{HCC}, CPTAC-PDA for CT pancreatic tumors ($n=46$)~\cite{CPTAC_PDA}, CPTAC-CCRCC for CT kidney tumors ($n=48$)~\cite{CPTAC_CCRCC}, and UCSF-PDGM for MR brain tumors ($n=494$)~\cite{UCSF_PDGM}. For each cohort, cases were included when imaging data, expert tumor contours, survival outcome information and the clinical variables required for the corresponding prognostic model were available.

\noindent\textbf{Automatic tumor segmentation and radiomics extraction.}
For each case, \methodname{} was used to generate an automatic tumor mask using only the target class-name prompt corresponding to the tumor type in that cohort, without manual spatial prompts or task-specific retraining. BiomedParse was applied as a competing text-driven automatic segmentation method under its corresponding inference setting. Expert manual tumor contours provided with each cohort were used as the reference masks for quantitative feature-agreement analysis and for constructing the expert-contour radiomics reference in prognostic modeling.

Radiomics features were extracted separately from \methodname{}-generated masks, BiomedParse-generated masks and expert manual contours using an identical preprocessing and feature-extraction pipeline within each cohort. The predefined feature panel included tumor burden and morphology-related descriptors, including volume, surface area, surface-to-volume ratio, maximum three-dimensional diameter, sphericity and elongation, together with selected first-order, grey-level co-occurrence matrix (GLCM) and grey-level run-length matrix (GLRLM) texture features. All image preprocessing, intensity normalisation, spatial resampling, feature discretisation and feature-extraction settings were held identical across the three mask sources within each cohort.

\noindent\textbf{Survival stratification using \methodname{}-derived features.}
To examine whether automatically extracted radiomics features retain prognostic information, patients within each cohort were divided into high- and low-feature groups using cohort-specific median cutoffs of representative \methodname{}-derived morphology and texture features. Survival distributions were estimated using the Kaplan-Meier method and compared using two-sided log-rank tests. Hazard ratios were estimated for the high-feature group relative to the low-feature group using univariable Cox proportional-hazards models. The Kaplan-Meier analyses in Fig.~\ref{fig:main_5}b are presented as cohort-level visualisations of representative morphology and texture features, whereas the multivariable prognostic modeling analysis in Fig.~\ref{fig:main_5}c provides the principal evaluation of downstream prognostic utility.

\noindent\textbf{Prognostic modeling using automatically derived radiomics.}
For each cohort, we evaluated four prognostic model settings: a clinical baseline model using the available clinical covariates only; a model combining clinical variables with radiomics features extracted from BiomedParse-generated automatic masks; a model combining clinical variables with radiomics features extracted from \methodname{}-generated automatic masks; and a model combining clinical variables with radiomics features extracted from expert manual tumor contours. The expert-contour radiomics model was reported as an expert-contour reference for the same prognostic modeling pipeline when manual delineations were available, rather than as a theoretical upper bound.

The same clinical covariates, candidate radiomics features, feature preprocessing procedures, model family and evaluation protocol were applied across the three radiomics mask sources within each cohort. Model discrimination was assessed using the concordance index (C-index). Statistical comparisons were performed between the clinical baseline model and the model augmented with \methodname{}-derived radiomics, and between the models augmented with BiomedParse-derived and \methodname{}-derived radiomics, corresponding to the comparisons shown in Fig.~\ref{fig:main_5}c. The panel displays two paired-bootstrap comparisons per cohort (clinical baseline versus clinical plus \methodname{} radiomics, and BiomedParse-derived versus \methodname{} radiomics). P values were adjusted across the 10 displayed cohort-level comparisons using the Benjamini--Hochberg procedure. The expert-contour radiomics result should be interpreted as a practical reference because prognostic performance depends not only on contour quality, but also on feature stability, cohort composition, available clinical covariates and model estimation.

\subsection{Clinician-guided refinement evaluation}

To evaluate whether \methodname{} supports efficient case-level correction when an automatically generated mask requires higher precision, we conducted a clinician-guided refinement analysis across three representative tumor segmentation tasks: liver cancer CT, acoustic neuroma MR and prostate cancer. This evaluation was designed to complement the retrospective cohort analysis above: whereas the retrospective analysis assessed scalable downstream quantitative utility, the refinement analysis assessed whether individual automatic predictions could be efficiently corrected through clinician interaction.

For each case, \methodname{} first generated an automatic tumor segmentation from a text prompt. Junior and senior doctors then reviewed the initial \methodname{} output and provided lightweight corrective interaction through the interactive refinement interface, after which the refined mask was generated using the cached decoder-logit-based refinement mechanism described above. As a comparison setting, junior and senior doctors provided bounding-box prompts for MedSAM on the same representative tasks. BiomedParse and automatic \methodname{} predictions were additionally included as automatic baseline settings.

Segmentation accuracy was evaluated using DSC, and practical interaction burden was evaluated using the latency required to produce the corresponding output. Accuracy-latency comparisons were reported for the average setting and for each representative tumor segmentation task (Fig.~\ref{fig:main_5}d). Representative refinement examples were further visualized by comparing the automatic \methodname{} mask, the refined \methodname{} mask, the MedSAM result based on a clinician-drawn bounding box, and the expert ground-truth contour (Fig.~\ref{fig:main_5}e; Extended Data Fig.~\ref{fig:extended_10}). This analysis was designed to evaluate segmentation correction efficiency and interaction burden, rather than diagnostic decisions, treatment recommendations or patient outcomes.

\section*{Data Availability}
Access to the in-house datasets from Sun Yat-sen Memorial Hospital, Sun Yat-sen University, is restricted. Inquiries regarding data access may be directed to the corresponding authors. The liver cancer dataset from Egypt is private due to hospital restrictions. The public SyReData release, including the released 2D images and masks, is available at \href{https://huggingface.co/datasets/McGregorW/SyReData}{https://huggingface.co/datasets/McGregorW/SyReData}; code and related resources are provided through the project repository at \href{https://github.com/xmed-lab/SyRe}{https://github.com/xmed-lab/SyRe}.

\section*{Code availability}
The source code is available at \href{https://github.com/xmed-lab/SyRe}{https://github.com/xmed-lab/SyRe}.

\section*{Acknowledgements}
This work was supported by grants from the National Natural Science Foundation of China (Grant No. 62306254), the Research Grants Council of the Hong Kong Special Administrative Region (Project Reference No. T45-401/22-N).
This work was also supported by the National Natural Science Foundation of China (Grant no. 82001768) and the Guangdong Basic and Applied Basic Research Foundation (Grant no. 2026A1515011134).

\section*{Author Contributions}
X.M.-L. designed the study. H.N.-W. developed and implemented a new machine-learning method, benchmarked machine-learning models, and analyzed model behavior. J.J.-M. and J.-S. collected the in-house data required for this study and performed the data labeling and the user study. 
L.H.-W., Q.X.-Z., Y.-Q., H.R.-L, and J.L.-L. implemented some baseline methods and experimental designs. 
M.-E provided the Egypt in-house data.
The clinical evaluation was organized by J.J.-M., while H.J.-H., B.X.-L., and W.F.-Q. provided support. 
H.N.-W. wrote the manuscript, with contributions from L.H.-W. and M.-E. and revisions by X.M.-L. 
All authors discussed the results and contributed to the final manuscript.

\section*{Competing Interests}
The authors declare no competing interests.

\clearpage
\bibliographystyle{unsrtnat}
\bibliography{bib,mybib}
\clearpage

\captionsetup[figure]{labelformat={default},labelsep=period,name={Extended Data Fig.}}
\setcounter{figure}{0}
\renewcommand{\theHfigure}{extended.\arabic{figure}}

\begin{figure}[t]
\centering
\includegraphics[width=1.0\linewidth]{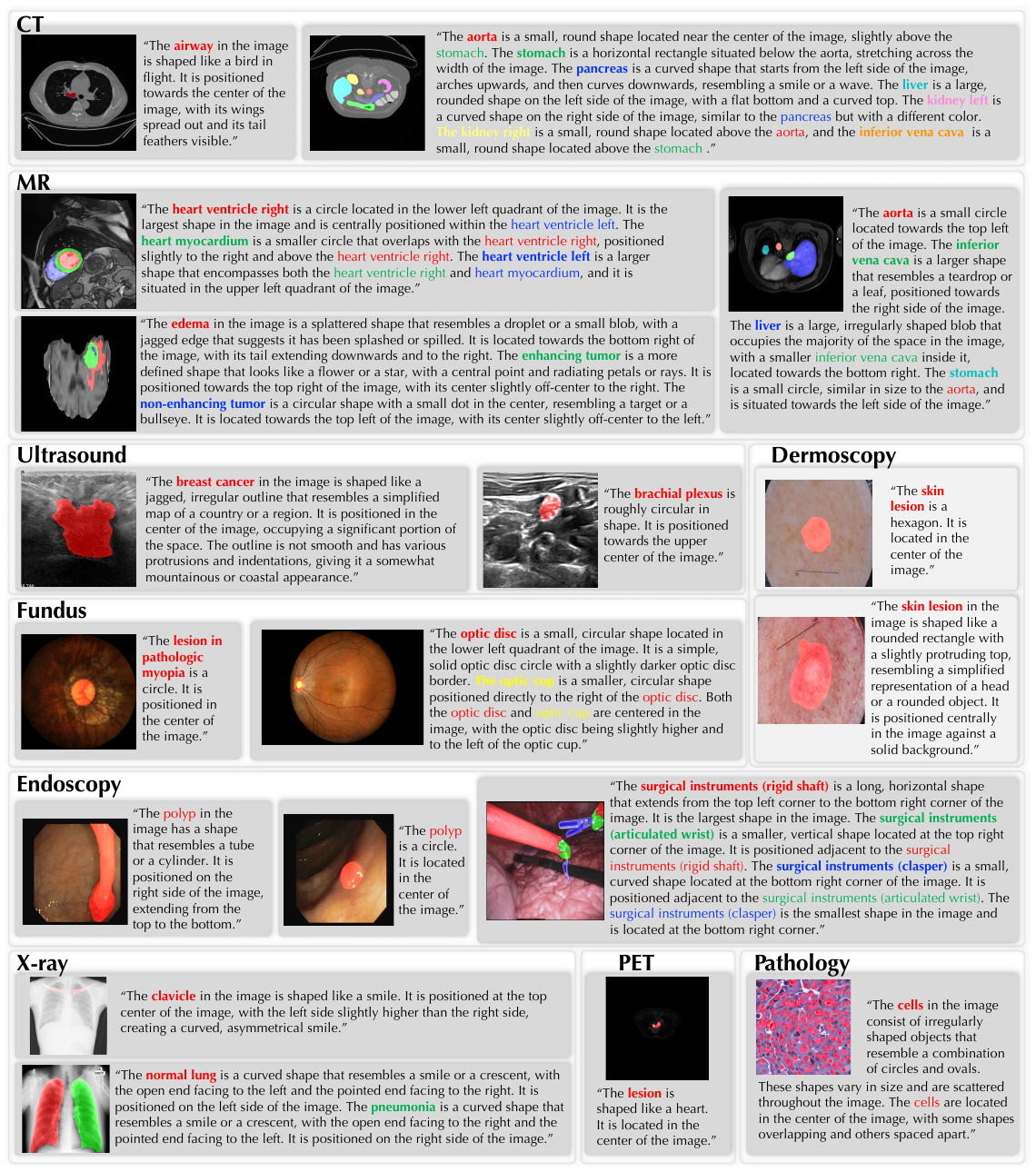}
\caption{Effectiveness of the Color Region Description (CRD) strategy. Examples across nine modalities show that the CRD strategy can produce comprehensive and accurate shape and relative location information. Trained on the generated image-mask-description triplets, \methodname{} established strong correlations between language and vision, leading to better handling of diverse morphological variants in disease samples of a specific category. 
}
\label{fig:extended_1}
\end{figure}

\begin{figure}[ht]
\centering

\includegraphics[width=\textwidth]{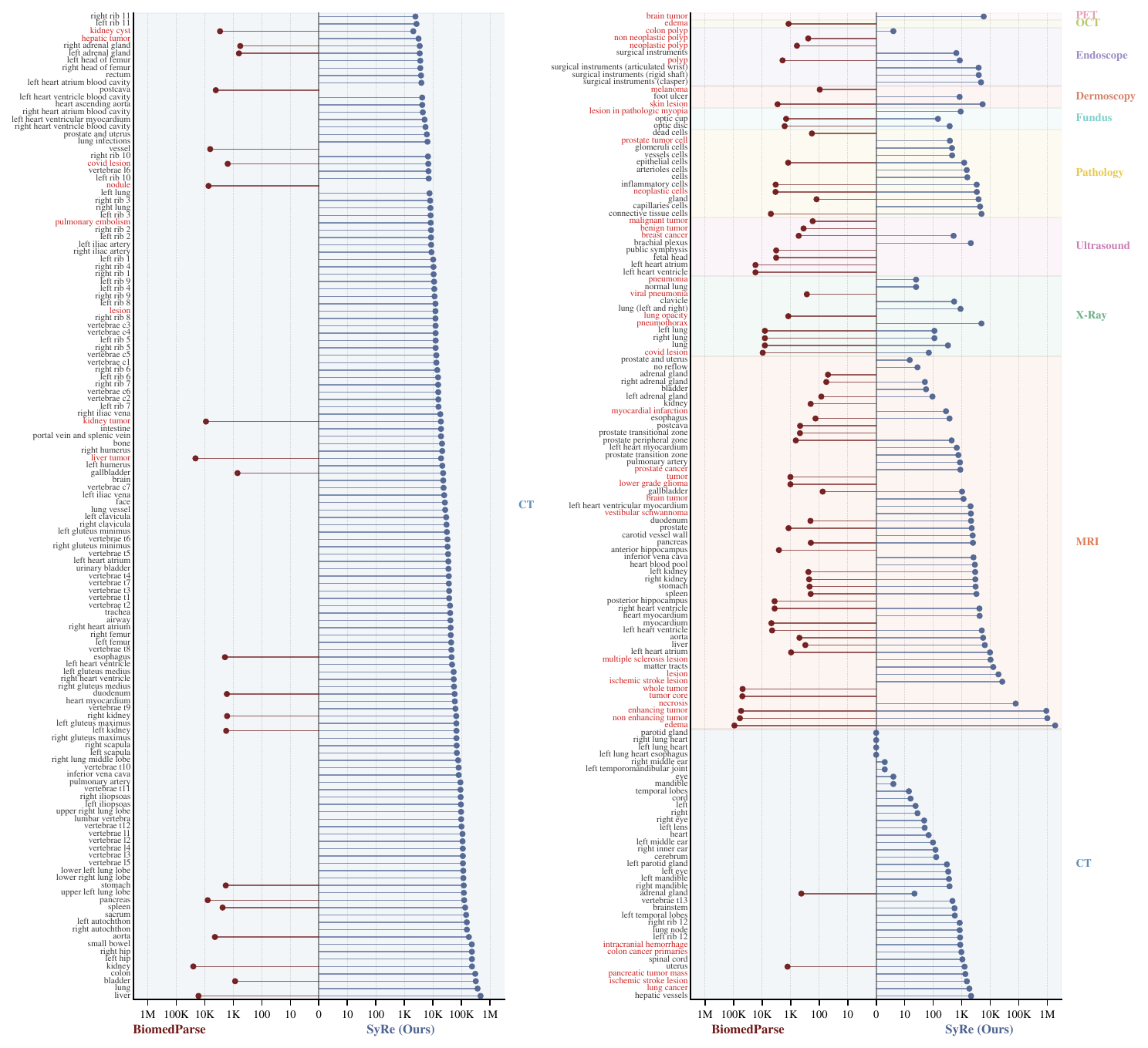}
\caption{\textbf{Comprehensive task coverage comparison between SyRe and BiomedParse
across all supported anatomical and pathological classes.}
Each row represents one class entry in the combined SyRe--BiomedParse coverage
table (269 total), grouped by imaging modality. The table combines the class
inventories used for the SyRe benchmark and the BiomedParseData comparison;
repeated class names can occur when the same class is represented in more than
one modality or benchmark entry, so 269 is not a count of unique classes.
(CT, MRI, X-Ray, Ultrasound, Pathology, Fundus, Dermoscopy, Endoscope, and PET).
The horizontal bars extend from the center axis in opposite directions: rightward bars
(\textcolor[HTML]{4e6390}{blue}) indicate the number of training masks available in SyRe,
and leftward bars (\textcolor[HTML]{6f1919}{dark red}) indicate the corresponding count in
BiomedParse.
Bar length is proportional to mask count on a $\log_{10}$ scale.
Classes highlighted in \textcolor{red}{red} correspond to pathological or
disease-related targets (e.g., tumors, lesions, infections).
SyRe was trained on 214 unique canonical classes spanning nine modalities, compared to 62
unique canonical classes in BiomedParse, with 173 classes exclusive to SyRe. Canonicalization
merges equivalent labels across source files (for example, formatting variants and side-specific
names) and applies predefined modality-specific aliases before counting. }
\label{fig:extended_2}
\end{figure}

\begin{figure}[ht]
\centering
\includegraphics[width=1.0\linewidth]{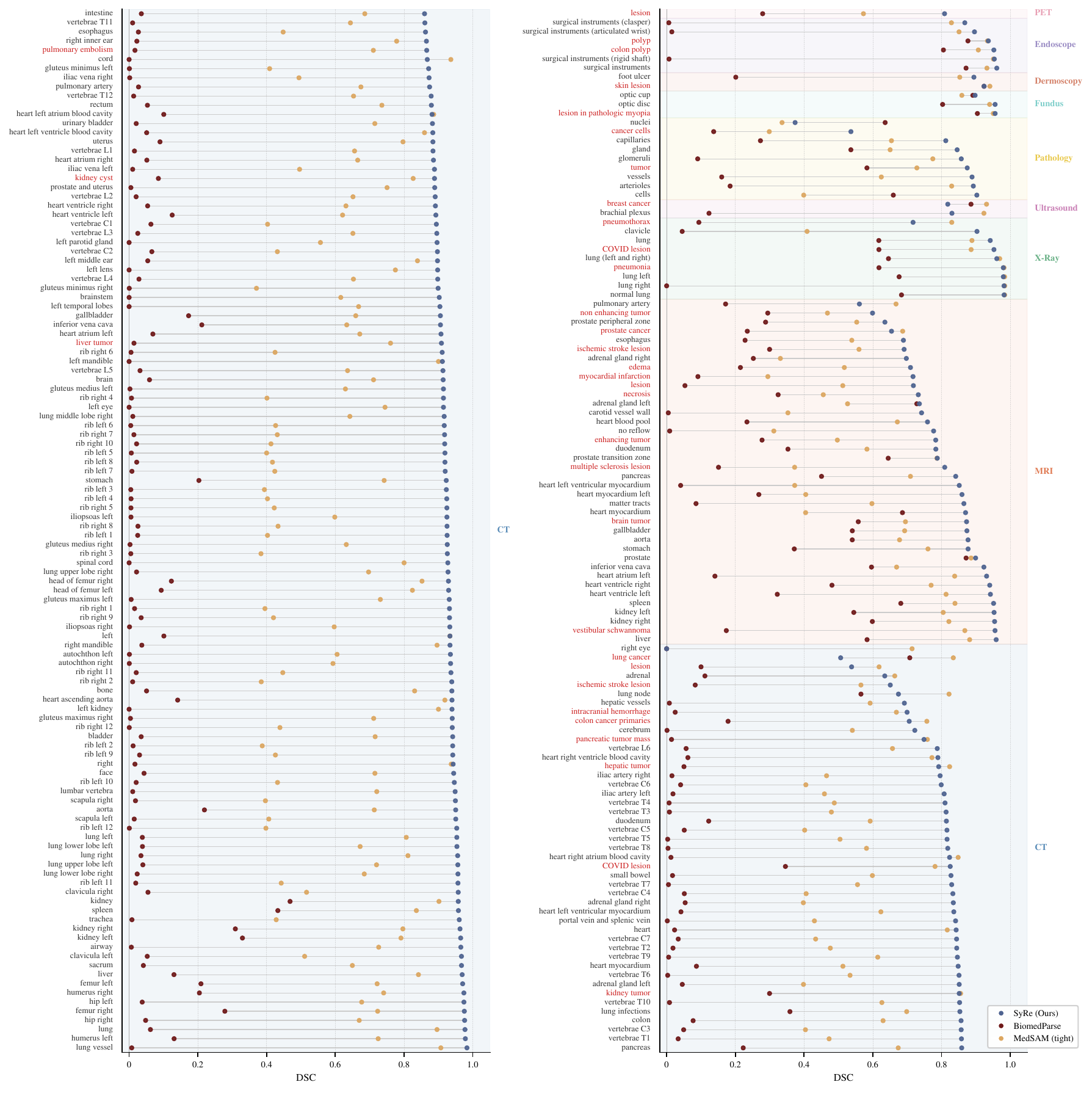}
\caption{\textbf{Per-class DSC comparison across 229 anatomical and pathological
categories on the internal test set.}
Each row represents one segmentation class, grouped by imaging modality (background
shading).
Dots indicate mean DSC for SyRe (blue), BiomedParse (red), and MedSAM with tight
bounding-box prompts (gold);
horizontal lines connect the three values to facilitate pairwise comparison.
Classes are ordered by SyRe DSC within each modality group (descending).
Disease-related classes (tumors, lesions, infections, \textit{etc.}) are labeled in red.
}
\label{fig:extended_3}
\end{figure}

\begin{figure}[t]
\centering
\includegraphics[width=1.0\linewidth]{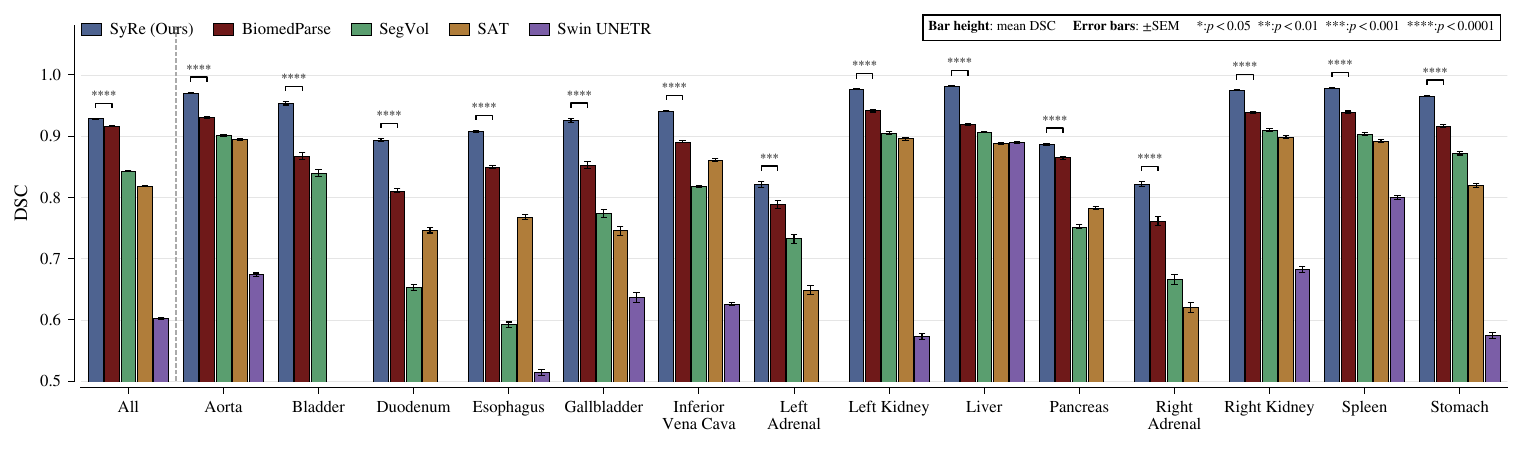}
\caption{
\textbf{Comparison of SyRe with BiomedParse and representative 3D segmentation models on a 3D CT organ-segmentation dataset.}
Bar plots compare SyRe with BiomedParse, SegVol, SAT, and Swin UNETR across 14 abdominal organs using the Dice similarity coefficient (DSC).
Bars indicate the mean DSC, and error bars indicate the standard error of the mean (SEM).
Statistical significance was assessed between the matched predictions of SyRe and BiomedParse using two-sided paired $t$-tests.
* indicates $p<0.05$, ** indicates $p<0.01$,
*** indicates $p<0.001$, and **** indicates $p<0.0001$.
}
\label{fig:extended_4}
\end{figure}

\begin{figure}[t]
\centering
\includegraphics[width=0.7\linewidth]{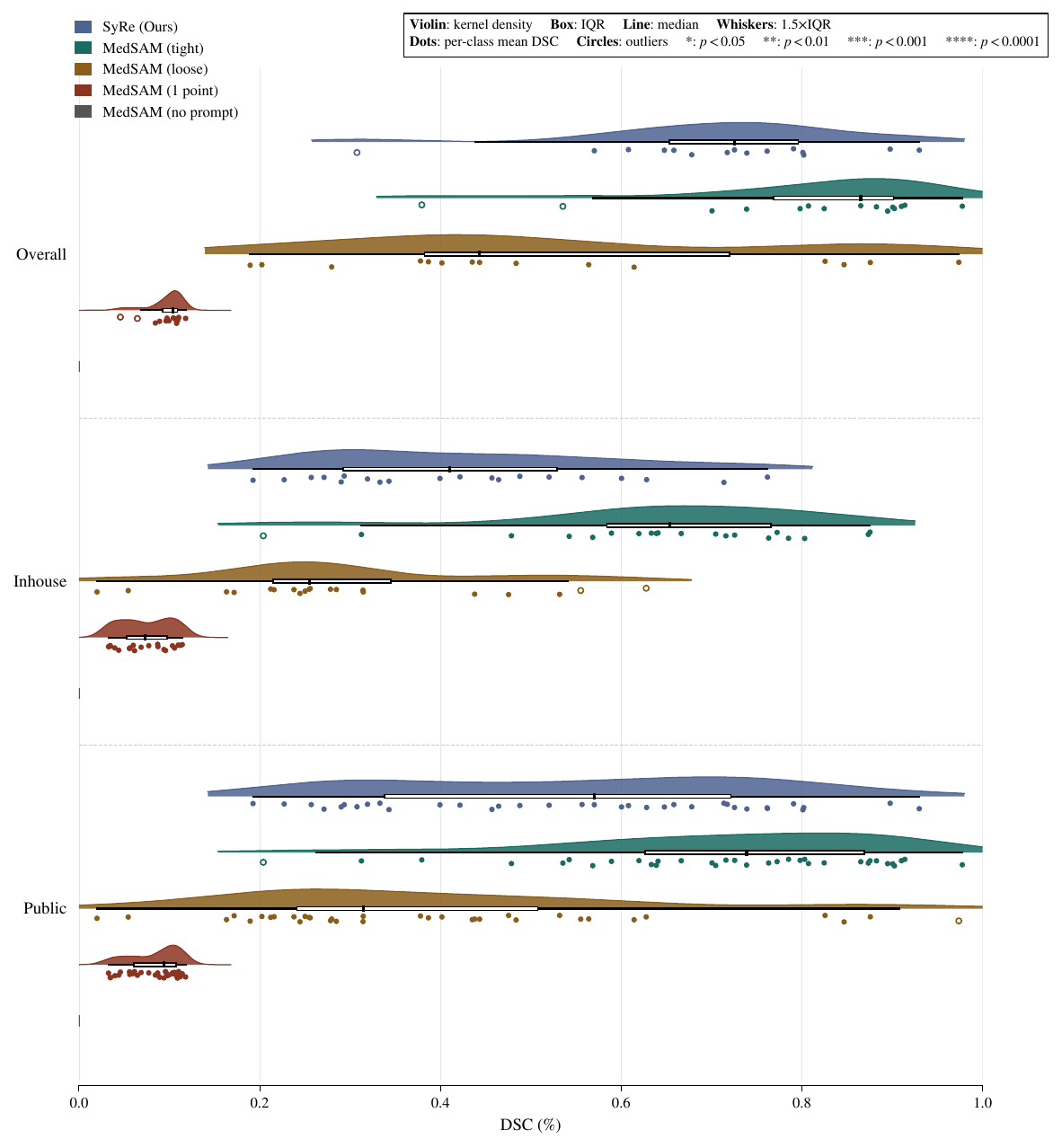}
\caption{\textbf{Distribution of per-class DSC across datasets for SyRe and MedSAM prompt
modes.}
Raincloud plots showing the distribution of per-class mean Dice Similarity Coefficient
(DSC)
for SyRe and four MedSAM prompt configurations (tight bounding box, loose bounding box,
single point, and no prompt) across three dataset groups: Inhouse ( 20-category cohort comprising 17 cancer types and 3 non-cancer diseases (cerebral haemorrhage, cerebral infarction and pulmonary embolism) ),
Public (external public benchmarks and AbdomenAtlas), and Overall (all datasets combined).
Each row shows a half-violin (kernel density estimate), interquartile box with median line,
1.5$\times$IQR whiskers, and per-class mean DSC as individual dots (outliers as open
circles).
}
\label{fig:extended_5}
\end{figure}

\begin{figure}[t]
\centering
\includegraphics[width=\linewidth]{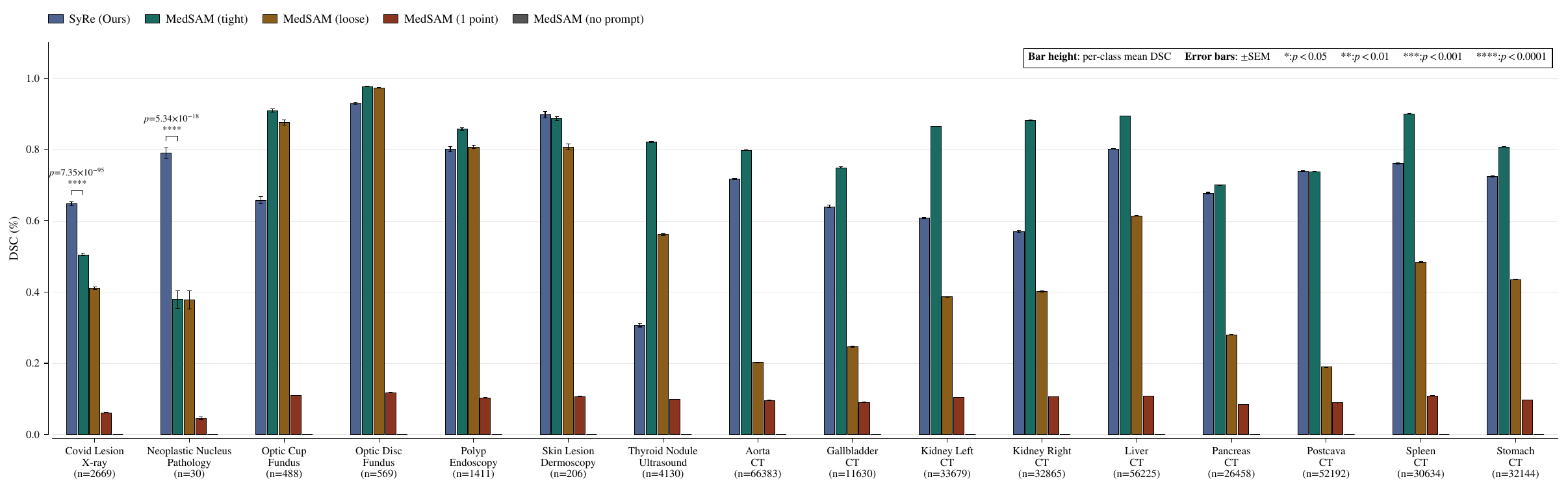}
\caption{ \textbf{Per-category DSC comparison on the 16 public external targets displayed in Fig.~\ref{fig:main_4}b.}
Category identities and displayed sample counts correspond to Fig.~\ref{fig:main_4}b. Bars show mean DSC with $\pm$SEM for SyRe and MedSAM under tight-box, loose-box, one-point and no-prompt conditions. }
\label{fig:extended_6}
\end{figure}

\begin{figure}[t]
\centering
\includegraphics[width=\linewidth]{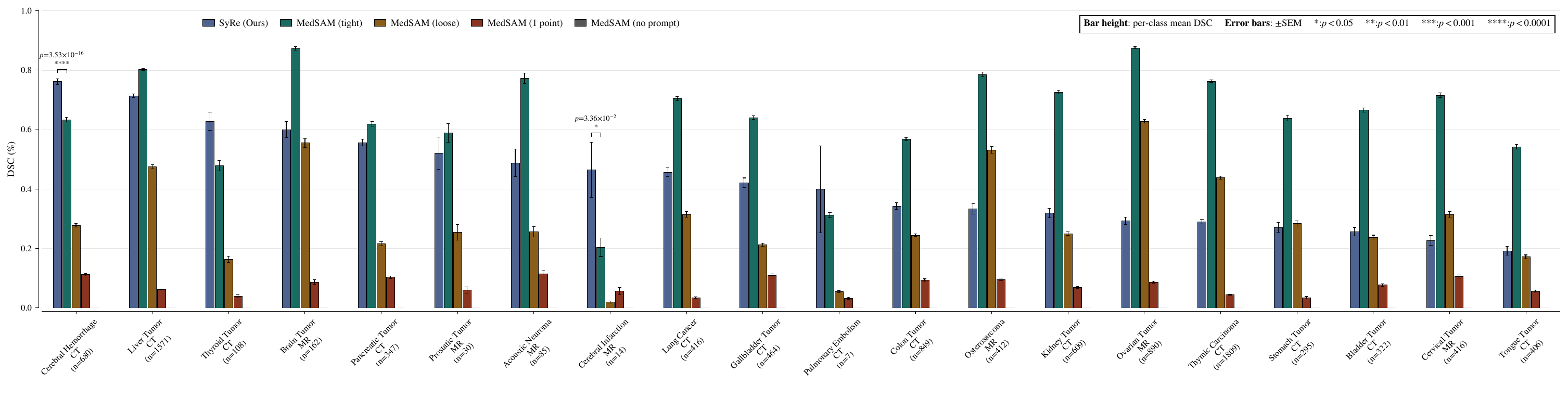}
\caption{ \textbf{Per-class DSC comparison of SyRe and MedSAM prompt modes on the 20-category
in-house cohort comprising 17 cancer types and 3 non-cancer diseases (cerebral haemorrhage, cerebral infarction and pulmonary embolism).}
Grouped bar charts showing mean DSC ($\pm$SEM) for each tumor class across SyRe and four
MedSAM prompt configurations (tight bounding box, loose bounding box, single point, and no
prompt). Classes are sorted by SyRe mean DSC in descending order, with sample size ($n$)
and
imaging modality indicated below each class label. All MedSAM prompt-mode results are
derived
from real evaluations on this cohort. }
\label{fig:extended_7}
\end{figure}

\begin{figure}[ht]
\centering
\includegraphics[width=1.0\linewidth]{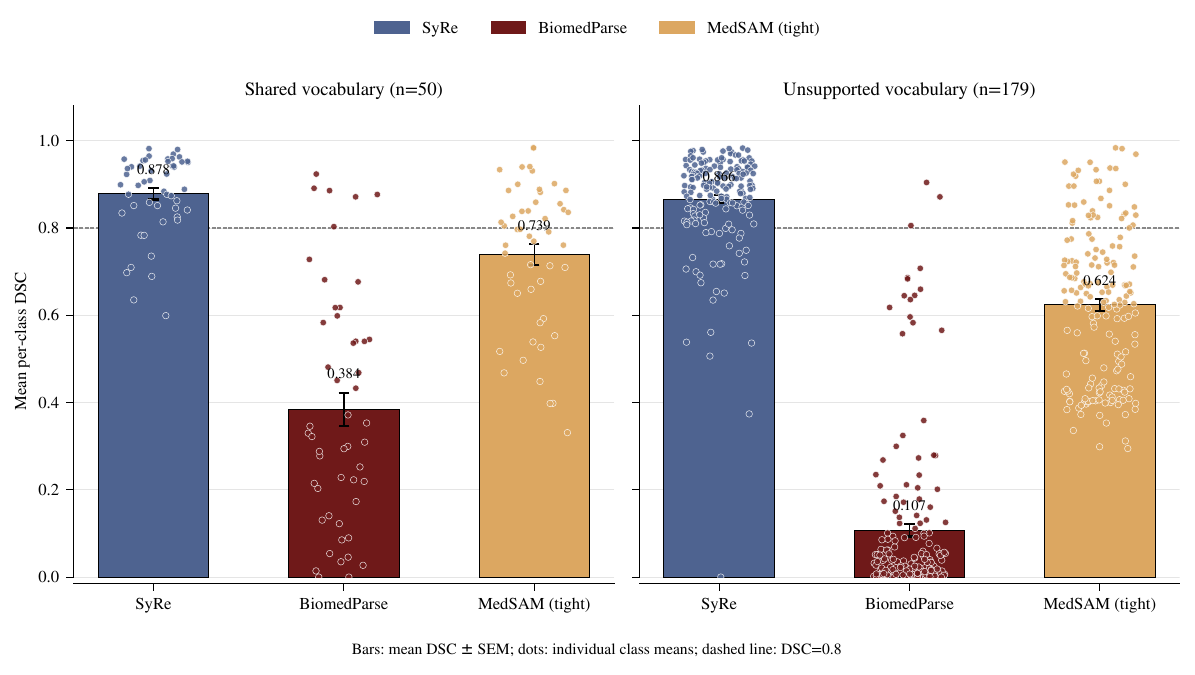}
\caption{\textbf{Vocabulary-aware class-wise comparison on the 229-class internal test set.}
The class-level results are split into entries shared with the BiomedParse inventory
($n=50$) and entries unsupported by that fixed vocabulary ($n=179$), using the same
modality- and canonical-class matching used for Extended Data Fig.~\ref{fig:extended_2}.
The shared-vocabulary subset provides the more direct comparison of segmentation
capability. Results for unsupported entries are reported separately because BiomedParse
cannot issue a target-specific prediction for those classes; their full-benchmark scores
therefore reflect vocabulary coverage as well as segmentation performance.}
\label{fig:extended_8}
\end{figure}

\begin{figure}[ht]
\centering
\includegraphics[width=1.0\linewidth]{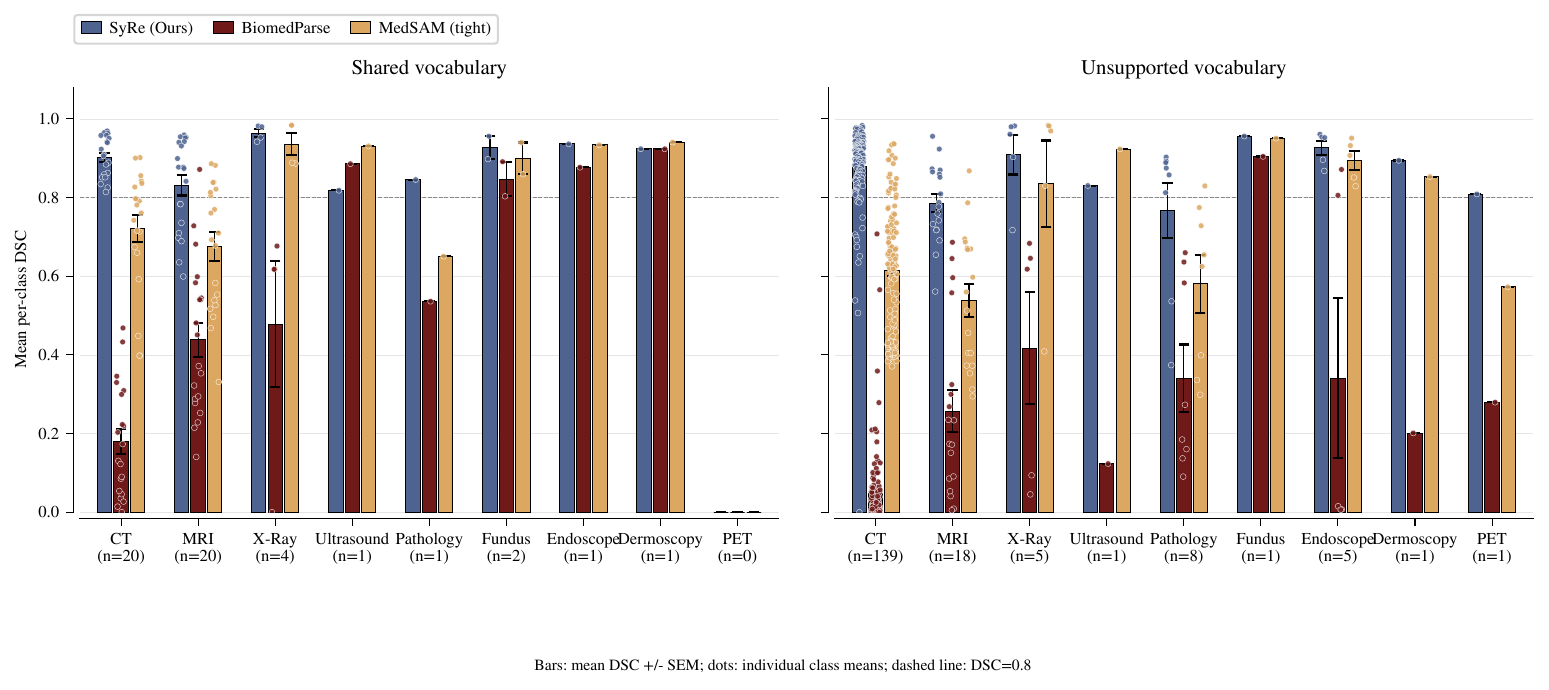}
\caption{\textbf{Vocabulary-aware comparison across imaging modalities.}
Bars show mean per-class DSC with SEM error bars for SyRe, BiomedParse and MedSAM
with tight bounding-box prompts. Dots show individual class means. Each panel reports
the number of class entries shared with the BiomedParse inventory and the number
unsupported by its fixed vocabulary. The dashed line marks DSC $=0.8$.}
\label{fig:extended_9}
\end{figure}

\begin{figure}[ht]
\centering

\includegraphics[width=0.89\linewidth]{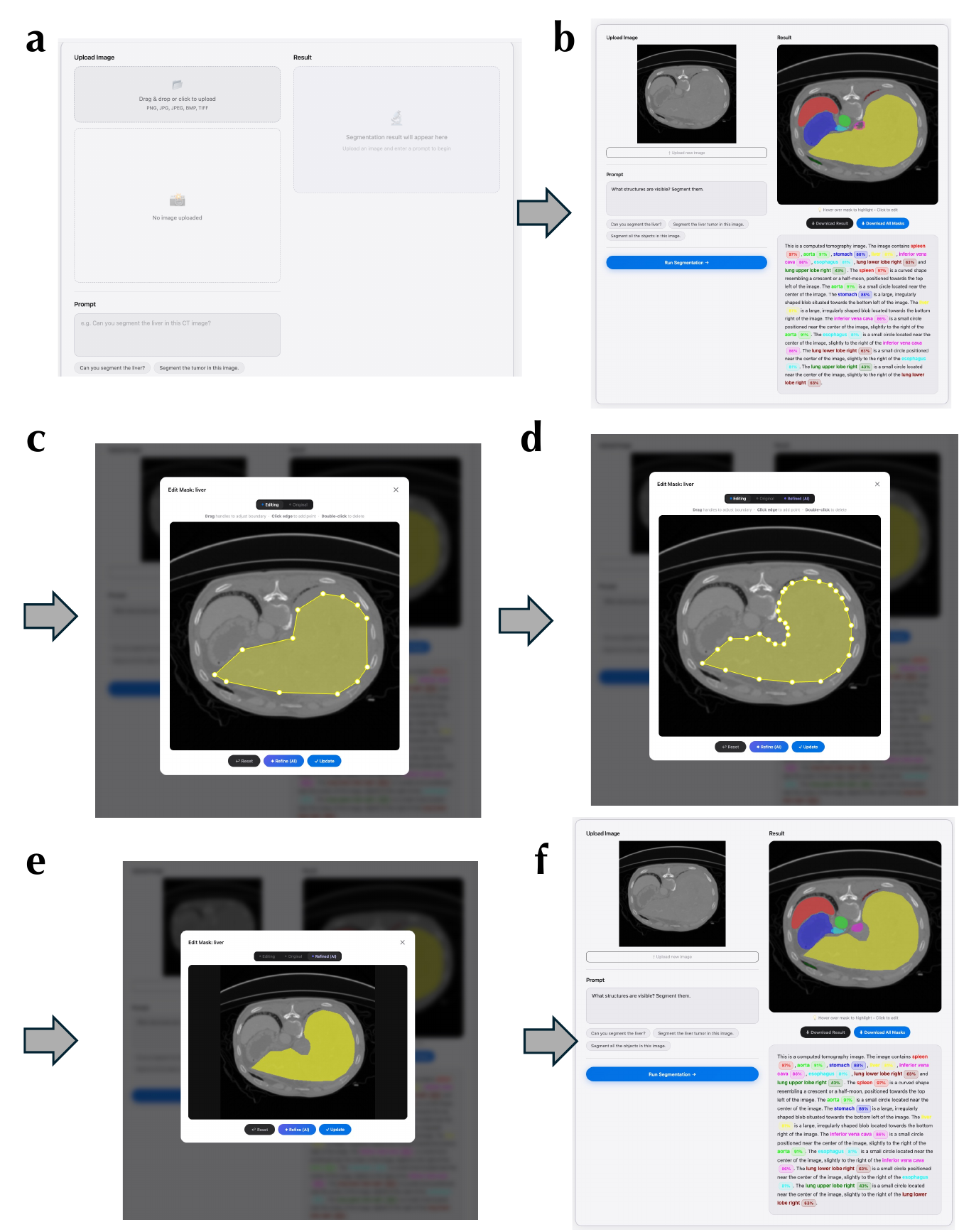}
\caption{
\textbf{Demo of the interactive refinement workflow.}
\textbf{a,} A user uploads a clinical image and enters a free-text prompt in the web interface. 
\textbf{b,} \methodname{} performs joint language generation and grounded segmentation in a single forward pass, displaying the predicted masks as semi-transparent overlays together with the textual response. 
\textbf{c,} The user opens the mask editor and selects a predicted mask for refinement. 
\textbf{d,} The mask contour is interactively adjusted through user-guided editing to correct under-segmented or over-segmented regions. 
\textbf{e,} The refined mask is updated in the editor and prepared for submission. 
\textbf{f,} The corrected result is returned to the main interface, where the refined mask replaces the original prediction, enabling iterative clinician-in-the-loop analysis without requiring a full re-inference of the model.
}
\label{fig:extended_10}
\end{figure}

\begin{figure}[!ht]
\centering
\includegraphics[width=1.0\linewidth]{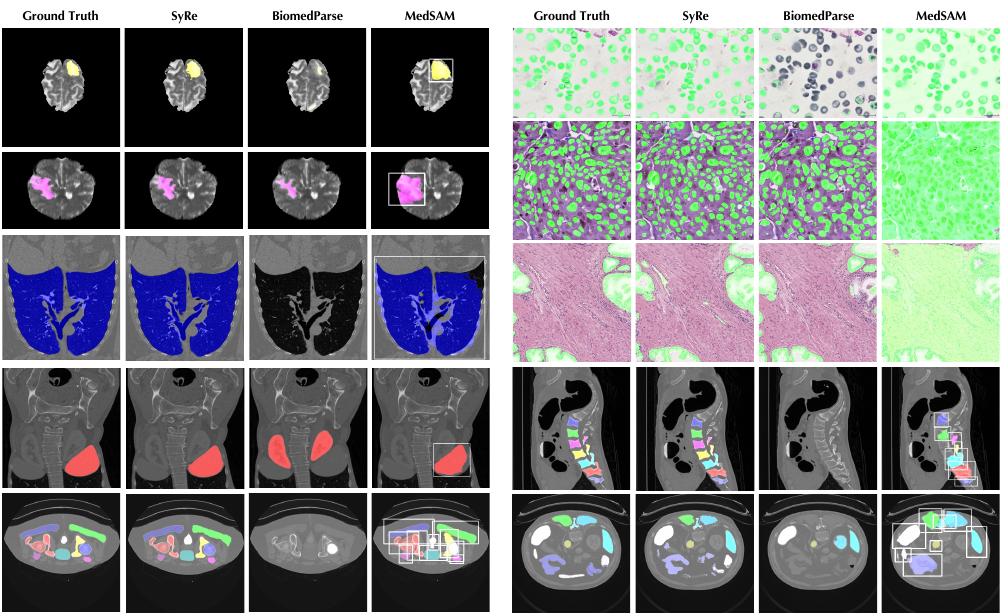}
\caption{Visual comparison samples from the held-out evaluation set. For both \methodname{} and BiomedParse, we utilize the class names of the segmentation targets as text prompts. For instance, we use "brain tumor" as the prompt for the first row. In the case of MedSAM, we employ tight bounding boxes as prompts, which are depicted as white boxes. 
}
\label{fig:extended_11}
\end{figure}

\clearpage
\end{document}